\documentclass[10pt,journal,compsoc]{IEEEtran}
\usepackage{amsmath,amsfonts}
\usepackage{algorithmic}
\usepackage{algorithm}
\usepackage{array}
\usepackage[caption=false,font=normalsize,labelfont=sf,textfont=sf]{subfig}
\usepackage{textcomp}
\usepackage{stfloats}
\usepackage{url}
\usepackage{verbatim}
\usepackage{graphicx}
\usepackage{cite}
\usepackage{hyperref}
\usepackage{subfiles}
\usepackage{booktabs}
\usepackage{xcolor}
\usepackage{multirow}
\usepackage{wrapfig}
\usepackage{colortbl}
\usepackage{makecell}
\usepackage{pifont}

\usepackage{enumitem}

\begin{document}



\title{RADAR: Revealing Asymmetric Development of Abilities in MLLM Pre-training}

 \author{
 Yunshuang Nie, Bingqian Lin, Minzhe Niu, Kun Xiang, Jianhua Han, Guowei Huang, Xingyue Quan, Hang Xu, Bokui Chen\IEEEauthorrefmark{2}, Xiaodan Liang\IEEEauthorrefmark{2}
 	\IEEEcompsocitemizethanks{
	\IEEEcompsocthanksitem 
	\IEEEauthorrefmark{2}Bokui Chen and Xiaodan Liang are the corresponding authors.\protect\\
    \IEEEcompsocthanksitem 
    Yunshuang Nie and Kun Xiang are with Shenzhen Campus of Sun Yat-sen University, Shenzhen, China. \protect\\
        E-mail:\{nieysh, xiangk\}@mail2.sysu.edu.cn.
            
        \IEEEcompsocthanksitem Xiaodan Liang is with Shenzhen Campus of Sun Yat-sen University, Shenzhen, China, Peng Cheng Laboratory, Guangdong Key Laboratory of Big Data Analysis and Processing, Guangzhou, 510006, China.
      \protect\\
      E-mail: liangxd9@mail.sysu.edu.cn.

     \IEEEcompsocthanksitem Bokui Chen is with Tsinghua Shenzhen International Graduate School, Tsinghua University, China. \protect\\
     E-mail: chenbk@tinghua.edu.cn.

    \IEEEcompsocthanksitem Bingqian Lin is with Shanghai Jiao Tong University, Shanghai, China.    \protect\\
    E-mail: linbq666@sjtu.edu.cn.

 \IEEEcompsocthanksitem Minzhe Niu, Jianhua Han, and Hang Xu are with Yinwang Intelligent Technology Co., Ltd. \protect\\
 E-mail: \{niuminzhe1@huawei.com, hanjianhua4@huawei.com, chromexbjxh@gmail.com\}

 \IEEEcompsocthanksitem Guowei Huang and Xingyue Quan are with Huawei's 2012 Lab. \protect\\
 E-mail: huangguowei@huawei.com, quanxingyue@huawei.com.



	}
		}

\markboth{Journal of \LaTeX\ Class Files,~Vol.~14, No.~8, August~2021}%
{Shell \MakeLowercase{\textit{et al.}}: A Sample Article Using IEEEtran.cls for IEEE Journals}

\IEEEpubid{0000--0000/00\$00.00~\copyright~2021 IEEE}

\maketitle

\begin{abstract}
Pre-trained Multi-modal Large Language Models (MLLMs) provide a knowledge-rich foundation for post-training by leveraging their inherent perception and reasoning capabilities to solve complex tasks. 
However, the lack of an efficient evaluation framework impedes the diagnosis of their performance bottlenecks.
Current evaluation primarily relies on testing after supervised fine-tuning, which introduces laborious additional training and autoregressive decoding costs.
Meanwhile, common pre-training metrics cannot quantify a model's perception and reasoning abilities in a disentangled manner, therefore hindering efficient enhancement for specific model abilities.
Furthermore, existing evaluation benchmarks are typically limited in scale or misaligned with pre-training objectives, which fail to provide accurate and direct assessment of multi-dimensional capabilities for pretrained MLLMs.
We propose \textbf{RADAR}, an efficient ability-centric evaluation framework for  \textbf{R}evealing \textbf{A}symmetric \textbf{D}evelopment of \textbf{A}bilities in MLLM p\textbf{R}e-training. RADAR involves two key components: (1) Soft Discrimination Score (SDS), a novel metric for robustly tracking ability development without fine-tuning,  based on quantifying nuanced gradations of the model preference for the correct answer over distractors; and (2) Multi-Modal Mixture Benchmark (M³-Bench), a new 15K+ sample benchmark for comprehensively evaluating pre-trained MLLMs' perception and reasoning abilities in a 0-shot manner. To construct M³-Bench, we unify authoritative benchmark datasets and carefully collect new datasets, extending the evaluation scope and addressing the critical gaps in current benchmarks.
Experimental results demonstrate a positive correlation between our introduced pre-training scores and fine-tuning performance.
With RADAR, we comprehensively reveal the asymmetric development of perceptual and reasoning capabilities in pretrained MLLMs across diverse factors, including data volume, model size, and pretraining strategy.
Our RADAR underscores the need for a decomposed perspective on pre-training ability bottlenecks, informing targeted interventions to advance MLLMs efficiently.
Our code is publicly
available at https://github.com/Nieysh/RADAR.

\begin{figure*}
\begin{centering}
\includegraphics[width=0.98\linewidth]{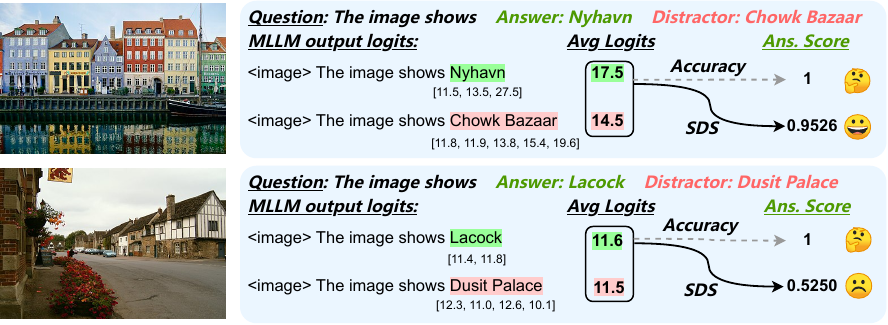}
\par\end{centering}
\caption{\textbf{The motivation of Soft Discrimination Score.} Accuracy employs the binary success criterion, which obscures the distinction between robust understanding and lucky guessing and is highly sensitive to near-correct distractor options. This is manifested as a tendency to overestimate the model's capabilities, leading to fluctuating results. In contrast, the soft discrimination score (SDS) can capture nuanced gradations of model preference for answers, reliably reflecting its understanding, and thus yielding stable test results.}
\label{fig:metric_motivation}
\end{figure*}

\section{Introduction} 
\IEEEPARstart{M}{ulti-modal} Large Language Models (MLLMs) serve as the critical foundation for a wide range of advanced AI applications~\cite{liu2023llava, li2024lavaonevision, 2025Qwen3-VL, wang2025internvl3_5, team2025gemma3, hurst2024gpt4o, 2025gemini3, Claude3, intelligence2025pi_0_5}. 
Current advanced MLLM training recipes typically follow a dual-stage process: pre-training and post-training. The pre-training stage has evolved from initially aligning modalities with limited data to now utilizing the larger portion of tokens among the two stages, enabling the model to progressively master perception and reasoning abilities across a diverse range of domains.

To develop the next generation of MLLMs, the pre-training phase is not merely a preliminary step but the foundational bottleneck that critically governs all subsequent capabilities~\cite{hurst2024gpt4o, 2025gemini3, comanici2025gemini2.5, liu2023llava, liu2024llavanext, li2024lavaonevision, bai2023qwenvl, wang2024qwen2vl,2025Qwen3-VL, chen2024internvl, chen2024internvl2, zhu2025internvl3, wang2025internvl3_5, mckinzie2024mm1, dai2024nvlm, agrawal2024pixtral, team2025gemma3, guo2025seed1_5, li2024baichuanomni}. A strong pre-trained MLLM, equipped with deep aligned representations across vision and language modalities and rich multi-domain knowledge, establishes a high-capacity latent space for eliciting complex instruction-following and reasoning behaviors, which is significant in post-training phases such as supervised fine-tuning (SFT) and reinforcement learning (RL).
Consequently, 
{\it rigorous and multifaceted evaluation of the pre-trained MLLMs is not an auxiliary activity but a core prerequisite}, which provides the essential diagnostics to quantify model's multi-dimensional capabilities, identifies bottlenecks in ability development from a decomposed perspective, and strategically guides the targeted interventions for certain abilities.

Conventional pre-training evaluation~\cite{lin2024vila, dai2024nvlm} relies on separate fine-tuning and costly autoregressive decoding of intermediate checkpoints.
This makes it prohibitively expensive to precisely track the continuous development of various abilities throughout pre-training and also creates a critical measurement gap, obscuring whether progress stems from improved core model capabilities or from later-stage optimizations.
An alternative tuning-free approach is in-context evaluation~\cite{mckinzie2024mm1, huang2025mir}, which uses few-shot examples to softly restrict the model's instruction-following behavior.
These evaluation paradigms predominantly rely on downstream, task-specific benchmarks, which conflate foundational ability acquisition in pre-training with instruction-following ability. Futhermore, they both introduce confounding factors (e.g., fine-tuning data, in-context example), hindering the assessment of model's true abilities.

Another tuning-free approach is to use pre-training metrics as proxies for pre-training quality, such as pre-training loss~\cite{vaswani2017transformer}, modality integration rate (MIR)~\cite{huang2025mir}, perplexity (PPL)~\cite{miaschi2021ppl}, and answer ranking accuracy~\cite{dai2023instructblip,li2024seedbench}.
However, these metrics are insufficient for precisely characterizing multi-dimensional abilities of pre-trained MLLMs. 
While pre-training loss and PPL offer an overall signal of pre-training quality, they are ineffective for comparing performance across distinct capability dimensions. Similarly, MIR, which assesses pre-training quality from the perspective of modality alignment, also lacks the granularity for quantifying different abilities in a disentangled manner.
Answer ranking strategy~\cite{brown2020gpt3, lin2022truthfulqa, dai2023instructblip, li2024seedbench}, which limits the model's vocabulary to multiple candidates and typically selects the option with the highest log-likelihood as the model's choice to calculate accuracy.
However, as illustrated in Figure~\ref{fig:metric_motivation}, the binary success/failure criterion of accuracy fails to distinguish between robust understanding and lucky guessing, and is highly susceptible to near-correct distractors (i.e., cases where the scores of the correct answer and a distractor are close), leading to unstable evaluation results.

Evaluation benchmarks are essential for assessing the multidimensional capabilities of MLLMs, providing a crucial foundation for iterative model improvement.
However, current benchmarks often require instruction-following capabilities and do not support direct evaluation of pretrained MLLMs due to their instability in instruction-following.
Furthermore, we roughly divide current evaluation benchmarks into general-purpose and domain-specific.
Many established general-purpose benchmarks, such as MME~\cite{fu2025mme}, MMBench~\cite{liu2024mmbench}, MM-Star~\cite{chen2024mmstar}, MM-Vet~\cite{Yu2024mmvet}, and BLINK~\cite{fu2024blink}, contain fewer than 5K question-answer (QA) pairs. The limited scale may introduce significant random error in the evaluation and pose a risk of substantial bias when assessing specific abilities.
In contrast, domain-specific benchmarks enable a more reliable assessment of targeted capabilities by providing sufficient test samples. Notable examples include MathVista~\cite{lu2024mathvista} for mathematics, Seephys~\cite{xiang2025seephys} for physics, and MMMU-Pro~\cite{yue2025mmmupro} for multi-disciplinary evaluation.
Some domain-specific benchmarks adopt specialized scenarios, including spatial reasoning benchmarks drawn from cognitive psychology~\cite{jia2025omnispatial}, embodied environments~\cite{du2024embspatial}, or 3D scenarios~\cite{yang2025vsi}. These benchmarks typically comprise challenging tasks tailored to specialized models, which require dedicated adaptation training to achieve strong performance. As pre-training data is typically broad but not domain-deep, general-purpose pre-trained MLLMs may struggle to achieve meaningful results on such benchmarks.
In summary, these existing benchmarks face notable challenges for evaluating pre-trained MLLMs, because they usually rely on limited data or misalign with the pre-training objective: requiring instruction-following rather than directly measuring a model’s pretrained ability or designing specialized scenarios beyond the capabilities of pretrained MLLMs.

In this paper, we propose an ability-centric evaluation framework, \textbf{RADAR}, aimed at \textbf{R}eve\textbf{A}ling \textbf{A}ymmetric \textbf{D}velopment of \textbf{A}bilities in MLLM p\textbf{R}e-training. By integrating a novel evaluation metric with a comprehensive benchmark, RADAR fulfills precise and stable tracking of multi-dimensional ability development during MLLM pre-training without additional fine-tuning and autoregressive decoding, which can efficiently identify performance bottlenecks and instruct future pre-training strategy design from a decomposed perspective.
First, we introduce the Soft Discrimination Score (SDS), a fine-tuning-free metric for robustly quantifying a model's multi-dimensional capabilities during pre-training. As shown in Figure~\ref{fig:metric_motivation}, given a multiple-choice question, SDS measures a model's relative preference for the correct answer by normalizing the average token-level logits of the correct answer with those of the distractors. This focus on relative preference, rather than binary correctness, makes SDS a robust and precise metric, as it captures nuanced gradations in the model's understanding as well as mitigating sensitivity to near-correct distractors.
Second, we propose the Multi-Modal Mixture Benchmark (M³-Bench), a large-scale benchmark of 15,894 samples designed to assess multi-domain perceptual and reasoning abilities during MLLM pre-training in a 0-shot manner. We employ a dual-strategy methodology for constructing M³-Bench: 1) systematically integrating and reformatting existing general-purpose and domain-specific resources to harness their diversity and expertise while reducing the requirement for instruction-following, and 2) extending the evaluation scope through meticulously collecting new datasets to address critical gaps in current benchmarks.
Specifically, the M³-Bench comprehensively evaluates 7 core task categories, structured into two principal abilities: \textbf{perception} (including {\it natural concept identification}, {\it cultural concept identification}, and {\it general visual question answering}) and \textbf{reasoning} (including {\it spatial reasoning}, {\it mathematical reasoning}, {\it physical reasoning}, and {\it multiple discipline visual question answering}).

Extensive pre-training experimental results consistently reveal a positive correlation between our introduced pre-training score 
and fine-tuning performance, confirming RADAR's effectiveness in indicating the pre-training quality.
Our evaluation involves three series of popular open-source pre-trained MLLMs, namely LLaVA-Onevision~\cite{li2024lavaonevision}, Qwen2-VL~\cite{wang2024qwen2vl}, InternVL-3.5~\cite{wang2025internvl3_5}, spanning from 0.5B to 14B. Evaluation results reveal a distinct performance dichotomy between perception and reasoning: 
The model with the largest pre-training dataset excels in perception, while another model with the largest model size achieves better reasoning performance.
Further analysis reveals an asymmetric development pattern: Increasing trainable parameters yields comparable gains in both perception and reasoning abilities. Increasing data reliably enhances perception across different model architectures, yet its benefits for reasoning plateau. Moreover, perception is effectively promoted by LLaVA-style alignment pre-training~\cite{liu2023llava}, whereas reasoning requires fully open, large-scale pre-training for significant gains. To investigate this divergence, we design controlled two-stage pre-training experiments to examine the impact of increasing the proportion of homogeneous data on different abilities. These experiments confirm that the reasoning bottleneck is not merely data volume. Such an asymmetric development pattern of perception and reasoning in pretrained MLLMs underscores the importance of a decomposed analysis of pre-training bottlenecks, guiding more efficient and targeted interventions for future MLLM advancement.

To summarize, our advantages are as follows:
\begin{itemize}

\item{\textbf{A Robust Evaluation Metric for Multi-dimensional Ability Measurement}: We propose the Soft Discrimination Score (SDS), which is a robust metric for evaluating multi-dimensional abilities of pre-trained MLLMs without fine-tuning.
SDS moves beyond binary accuracy to measure a model's nuanced preference for correct answers via normalized token-level logits, thereby providing a precise and stable signal for tracking multi-dimensional ability progression.}

\item{\textbf{A Comprehensive Benchmark for Core Perception and Reasoning Ability Assessment}: We construct M³-Bench, a large-scale benchmark comprising 15K+ questions for perception and reasoning. It systematically unifies and reformats general-purpose and domain-specific resources while extending evaluation scope with newly collected data, enabling 0-shot comprehensive evaluation tailored for pretrained MLLMs.}

\item{\textbf{An Empirical Analysis Revealing Asymmetric Development}: Through extensive evaluation, we identify and analyze a fundamental dichotomy: perception and reasoning abilities scale differently with data and parameters. We find that perception benefits consistently from increased data and alignment-style pre-training, while reasoning exhibits diminishing returns from data scaling and requires large-scale pre-training with more trainable parameters for substantial gains. Controlled experiments confirm that the reasoning bottleneck is not solely a data-volume issue.}

\end{itemize}

\begin{figure*}
\begin{centering}
\includegraphics[width=0.98\linewidth]{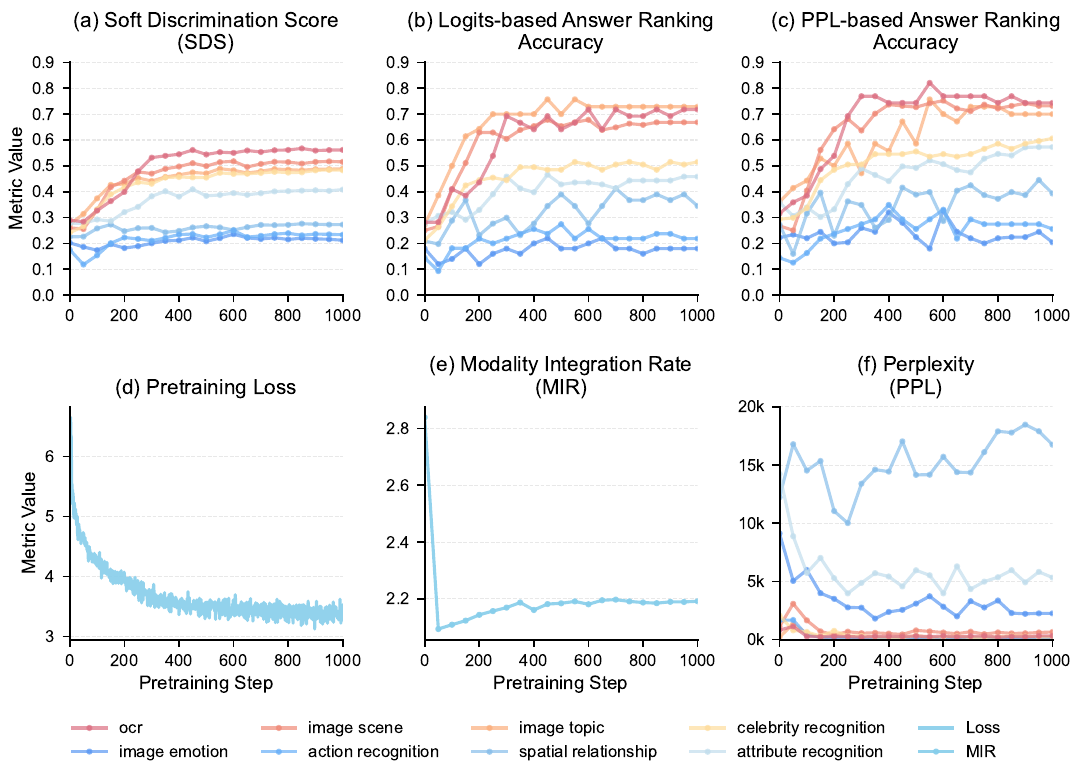}
\par\end{centering}

\vspace{-0.2cm}
\caption{\textbf{Comparison between Soft Discrimination Score and existing pre-training metrics on 8 perception-related L3 category tasks in MMBench. } We conduct the LLaVA-style pre-training with Qwen2-0.5B-Instruct~\cite{yang2024qwen2} model and SigLIP-so400m-patch14-384~\cite{zhai2023siglip} on 300K general pre-training data collected from web. We save the checkpoint every 50 steps for assessment. We evaluate the checkpoints on 20 L3 category tasks in MMBench dev (en) split~\cite{liu2024mmbench}, e.g., {\it image scene}. For SDS, PPL, and answer ranking-based accuracy, we assess each checkpoint on repurposed MMBench (with its original number of distractors). The logits-based answer ranking uses the option with the highest average token logits as the model's choice. Since PPL is not directly comparable across different task corpora, we derive a PPL-based answer-ranking strategy to compute accuracy.
For pre-training loss, we use the value computed with the pre-training data. For MIR, we use the same dataset as in~\cite{huang2025mir} because it is a task-agnostic metric.
For clarity, we only present 8 perception-related L3 category tasks in MMBench here. Comparisons across the reasoning-related L3 category tasks are provided in the supplementary material.
}
\label{fig:metric_comparison}
\end{figure*}

\section{Related Work}

\subsection{Pre-training Paradigm of MLLM}
The pre-training paradigm for Multi-modal Large Language Models (MLLMs) has undergone significant changes in objectives~\cite{liu2023llava,liu2024llavanext,li2024lavaonevision,mckinzie2024mm1,dai2024nvlm,bai2023qwenvl,wang2024qwen2vl,2025Qwen3-VL,chen2024internvl,chen2024internvl2,zhu2025internvl3,wang2025internvl3_5,team2025gemma3}. Early approaches, such as the initial LLaVA~\cite{liu2023llava} series, adopt a simplified vision-language projector pre-training to achieve basic feature alignment. The core objective is to efficiently bridge the modality gap. However, recent models have redefined pre-training as a more critical stage for learning diverse knowledge. For instance, the Qwen-VL~\cite{2025Qwen3-VL} and LLaVA-OneVision~\cite{li2024lavaonevision} series further divide pre-training into alignment and broad knowledge-acquisition phases, emphasizing the importance of extensive multi-modal knowledge learning by scaling up pre-training data and unlocking all parameters. Furthermore, InternVL3.5~\cite{wang2025internvl3_5} series integrates native multi-modal pre-training by mixing pure text and image-text data to simultaneously develop linguistic and cross-modal capabilities.

This evolution, from a simple alignment step to a comprehensive, data-centric, and large-scale pre-training process, aims not only to integrate modalities but also to build a generalizable multi-modal foundation, making the pre-training stage a decisive factor for achieving strong downstream performance across diverse and challenging benchmarks.


\subsection{Evaluation Benchmark for MLLM}
Existing benchmarks for evaluating foundational capabilities can be roughly categorized into general-purpose datasets for comprehensive assessment and domain-specific datasets for specialized evaluation.
Popular general benchmarks such as MMBench~\cite{liu2024mmbench}, MMStar~\cite{chen2024mmstar}, MM-Vet~\cite{Yu2024mmvet}, BLINK~\cite{fu2024blink}, and MME~\cite{fu2025mme} typically contain fewer than 5K image-related test samples, assessing an MLLM's capabilities from various perspectives only at a small scale.
As models advance in handling complex tasks, domain-specific datasets have emerged to assess model capabilities in specialized fields, such as mathematical reasoning~\cite{lu2024mathvista, wang2024mathvision,zhang2024mathverse, qiao2025wemath, zou2025dynamath}, physical reasoning~\cite{xiang2025seephys, shen2025phyx, wang2025physunibench}, multidisciplinary knowledge understanding and reasoning~\cite{yue2024mmmu, yue2025mmmupro}, spatial reasoning~\cite{liu2023visualspatial, kamath2023wahtsup, cheng2024spatialrgpt, du2024embspatial, yang2025vsi, jia2025omnispatial}. These benchmarks typically consist of sufficient quantities of domain-specific image-guided questions.
However, certain benchmarks are designed to test specialized models, which are often beyond the abilities of general pre-trained MLLMs. This misalignment of evaluation objective makes them unsuitable for evaluating foundational core abilities of pre-trained models. For instance, when assessing spatial reasoning capabilities, using benchmarks like OmniSpatial~\cite{jia2025omnispatial} or VSI~\cite{yang2025vsi} to test a pretrained MLLM yields results around random chance.
Furthermore, these benchmarks conflate foundational ability acquisition in pre-training with instruction-following ability and require costly autoregressive decoding process, hindering the efficient evaluation of the true capabilities of pre-trained MLLMs.

Unlike prior efforts, the primary advantage of our introduced M³-Bench lies in its compatibility with our proposed metric to evaluate pre-trained MLLMs by bypassing autoregressive generation and instruction-following requirements. The M³-Bench leverages the diversity and expertise of existing general-purpose and domain-specific benchmark resources while collecting new datasets to address critical gaps in current benchmarks, enabling comprehensive 0-shot evaluation of pre-trained MLLMs' perception and reasoning abilities.

\subsection{Pre-training Evaluation for MLLMs}
Several systematic studies have explored the impact of pre-training (including both textual and multi-modal pre-training) on MLLMs~\cite{dai2024nvlm, lin2024vila, mckinzie2024mm1, han2025learningtoseebeforeseeing, huang2025mir}.  
Previous works~\cite{lin2024vila, mckinzie2024mm1, dai2024nvlm} mainly explore the effects of model structure, data mixing, and which parameters to freeze in pre-training, and then test the fine-tuned model or conduct in-context evaluation for analysis.
A different perspective introduces Modality Integration Rate (MIR)~\cite{huang2025mir}, which is an effective metric that evaluates MLLM pre-training quality from a distribution distance perspective.
Besides multi-modal pre-training, a recent study~\cite{han2025learningtoseebeforeseeing} presents the first systematic investigation into the origin of visual priors in large model language pre-training, identifying that visual priors can be decomposed into perceptual and reasoning components. Based on the distinct sources and scaling patterns of these components, they further propose a data-centric methodology for training vision-aware LLMs.

Inspired by prior work, our study designs a robust evaluation metric and a comprehensive diagnostic benchmark to track the development of multi-dimensional abilities for pre-trained MLLMs without finetuning. Through extensive evaluation with RADAR, we reveal the fundamental asymmetric development of perception and reasoning, identifying ability bottlenecks from a decomposed perspective that can inform targeted strategies for future MLLM advancement.

\section{Method}
\label{Method}
In this section, we first review common pre-training metrics and detail the formulation of our proposed Soft Discrimination Score (SDS). Then we compare its efficacy against common pre-training metrics, such as loss, perplexity (PPL), modality integration rate (MIR), and answer ranking accuracy, in tracking ability development during MLLM pre-training (Sec.~\ref{sec:Soft Discrimination Score}). Second, we present the construction of M³-Bench, which employs a dual methodology: 1) unifying and reformatting existing general-purpose and domain-specific benchmarks to establish a diverse assessment foundation for 0-shot evaluation, and 2) augmenting this foundation with meticulously collected samples from Wikipedia and the web to extend the evaluation scope and address the gaps in current benchmarks (Sec.~\ref{sec:Multi-Modal Mixture Benchmark}).

\subsection{Soft Discrimination Score}
\label{sec:Soft Discrimination Score}

Before formally introducing the formulation of our metric, we first review the current pre-training metrics.

A commonly used metric for MLLMs, pre-training loss, is usually based on the autoregressive objective,
which is computed as the negative log-likelihood over all tokens in the pre-training label sequence~\cite{vaswani2017transformer}.
Another widely used metric is perplexity (PPL)~\cite{miaschi2021ppl}, which is computed with the exponentiated average negative log-likelihood of the pre-training label.
While the pre-training loss and its derived metric, PPL, provide an overall measure of a model's fit to its training data distribution, they exhibit significant limitations when used as primary metrics for evaluating multi-dimensional capabilities in pre-trained MLLMs, because their numerical values are not directly comparable across different model architectures or different task corpora.
Modality integration rate (MIR)~\cite{huang2025mir} measures the pre-training modality alignment quality of MLLMs with Fr\'{e}chet Inception Distance~\cite{heusel2017gans}.
While providing a novel perspective of pre-training evaluation, MIR shares critical limitations with loss and perplexity as evaluation tools for model capabilities, because it is generally agnostic to different types of inputs and therefore cannot discriminate between capability types.

Besides the previous metrics, a typical indicator that can distinguish different abilities is accuracy~\cite{li2024seedbench}. In order to conduct 0-shot evaluation, the frequently used approach is the answer ranking strategy~\cite{dai2023instructblip, li2024seedbench}, which is usually implemented with multiple-choice questions. Based on the model's prediction of the answer and the distractors, the option that has the highest score is considered the model's choice. Specifically, define a pre-trained MLLM probability distribution as $P$, the input image and question as $I$ and $Q$, respectively. For the $i$-th test sample, there is a set of candidate answers $C=\{c_1,c_2,...,c_M\}$, and the model's choice $y$ is the candidate that receives the highest likelihood score:
\begin{align}
y &= \mathop{\arg\max}_{c\in C}P(c|Q,I).
\end{align}
We omit the subscript $i$ to maintain clarity.
The overall accuracy is the mean of the indicator function across all $N$ test samples:
\begin{align}
\mathrm{Accuracy} &= \frac{1}{N}\sum_{i=1}^{N}1(y=\mathrm{A}).
\end{align}
where $\mathrm{A}$ is the ground-truth answer for the sample, and $1(\cdot)$ is the indicator function (outputs 1 if the condition is true, else 0).

As a binary metric, accuracy provides no insight into why the model succeeds or fails, obscuring the distinction between robust understanding and lucky guessing, leading to volatile and less interpretable performance measurements.

To address these limitations and enable robust, precise tracking of multi-dimensional ability development during MLLM pre-training, we propose the Soft Discrimination Score (SDS).
Like the answer ranking strategy, SDS also relies on multiple-choice questions to ensure comparability across different samples and limits the model's vocabulary to multiple candidates to bypass autoregressive decoding and enable 0-shot evaluation. However, instead of using the binary success criterion, SDS measures the model's relative preference for the correct answer by normalizing the average token-level logits of correct answers with those of distractors. Averaging logits at the token level provides a fine-grained view, and normalization ensures consistency across different questions. This design of relative preference makes SDS a robust and precise metric, as it captures nuanced gradations in the model's understanding while mitigating sensitivity to near-correct distractors.

Denote a pre-trained MLLM as $F$, which outputs logits for each token in the tokenizer vocabulary under specific conditions. We compute the average token logits $s_{c_j}$ for each candidate $c_j$ independently:
\begin{align}
s_{c_j} &= \frac{\sum_{t=1}^{T}F(c_{j,t}|c_{j,<t},Q,I)}{T}.
\end{align}
where $T$ denotes the token length of $c_j$. Then we compute the normalized score $p$ for candidates in $i$-th test sample:
\begin{align}
p &= \frac{\mathrm{exp}(s_{c_{j}})}{\sum_{j=1}^{M}\mathrm{exp}(s_{c_{j}})}.
\end{align}
where $M$ is the number of candidates $C$.
We extract the normalized score of the accurate answer $p_{A}$ as the score for the model on the $i$-th test sample.
Finally, we compute the average score on all samples as the SDS score:
\begin{align}
\mathrm{SDS} = \frac{1}N{\sum_{i=1}^{N}p_{A}}
\end{align}


To demonstrate the effectiveness of SDS in tracking the development of multi-dimensional capabilities, we compare it with common metrics in pre-training evaluation.
We conduct evaluation on L3 category tasks in perception and reasoning on the dev english split of MMBench~\cite{liu2024mmbench}, providing a shared basis for comparison.
As shown in Figure~\ref{fig:metric_comparison}, SDS (Figure~\ref{fig:metric_comparison} (a)) can quantify the progress across different ability dimensions while also capturing their saturation (similar to the convergence observed in pre-training loss, which generally occurs between 400 and 600 steps). In contrast, overall metrics, like pre-training loss (Figure~\ref{fig:metric_comparison} (d)) and MIR (Figure~\ref{fig:metric_comparison} (e)), lack this dimensional granularity.
For comparison, we implement two answer-ranking strategies: 1) logits-based answer ranking accuracy (Figure~\ref{fig:metric_comparison} (b)), which selects the option with the highest average token logits as the model's choice, and 2) PPL-based answer ranking accuracy (Figure~\ref{fig:metric_comparison} (c)), which selects the option with the lowest PPL as the model's choice, since raw PPL (Figure~\ref{fig:metric_comparison} (f)) values vary widely across tasks and are not directly comparable. While both accuracy variants reveal differences in ability, their trajectories exhibit irregular fluctuations and obscured saturation, complicating analysis. In comparison, SDS clearly reflects gradual improvement in capability, offering more interpretable and stable signals for monitoring pre-training progress.
Overall, SDS provides a stable, fine-grained, and interpretable metric that directly tracks the development of multi-dimensional capabilities throughout the pre-training process.

\subsection{Multi-Modal Mixture Benchmark}
\label{sec:Multi-Modal Mixture Benchmark}
Current benchmarks face key challenges for pre-training evaluation: limited scale and misalignment with the pre-training objective. To address this, we introduce a new multi-modal evaluation benchmark designed for pre-trained MLLMs, the Multi-modal Mixture Benchmark (M³-Bench), prioritizing large-scale, diverse 0-shot evaluation of core abilities directly relevant to foundational model pre-training.

We ground our benchmark in the established understanding that effective MLLM pre-training aims to forge two interconnected yet distinct core competencies: \textbf{perception}, which involves grounding language in visual signals, and \textbf{reasoning}, which requires deriving inferences from this grounded understanding. This dichotomy is a central tenet of MLLM development and evaluation, as reflected in seminal benchmarks such as MME~\cite{fu2025mme} and MMBench~\cite{liu2024mmbench}, which systematically decouple and probe these capabilities. Informed by the focus areas of cutting-edge MLLM research~\cite{wang2025internvl3_5,2025Qwen3-VL,guo2025seed1_5, team2025gemma3}, we select and define 7 specific task categories that reflect these critical capabilities.

For perception, we move beyond elementary object recognition to evaluate more fine-grained visual understanding. Our designed tasks target the following specific capabilities:
\begin{itemize}
    \item Natural Concept Identification~\cite{hu2023oven}: This task evaluates fine-grained recognition and attribute disambiguation of natural entities, testing a model's breadth and precision of visual knowledge.
    \item Cultural Concept Identification~\cite{hu2023oven}: This task tests the recognition of socially and culturally grounded concepts, probing the deep alignment between visual perception and real-world knowledge.
    \item General Visual Question Answering~\cite{liu2024mmbench, Yu2024mmvet, chen2024mmstar, cheng2025simplevqa, fu2025mme, zhang2024mmerealworld}: Serving as an integrative test, this task incorporates diverse questions requiring holistic image interpretation. It measures the model's overall capacity for open-ended visual comprehension.
\end{itemize}

For reasoning, we focus on inference capabilities essential for complex problem solving:
\begin{itemize}
    \item Spatial Reasoning~\cite{kamath2023wahtsup, du2024embspatial, liu2023visualspatial}: We evaluate the understanding of spatial relationships, object layouts, and orientations using diagrams and scene imagery. This capability is a prerequisite for complex interaction downstream tasks.
    \item Mathematical Reasoning~\cite{qiao2025wemath, lu2024mathvista, wang2024mathvision, zou2025dynamath, zhang2024mathverse}: This task evaluates quantitative and symbolic logic by requiring models to parse visually presented data (e.g., charts, diagrams, equations) and perform necessary calculations.
    \item Physical Reasoning~\cite{xiang2025seephys, wang2025physunibench, shen2025phyx}: To test deep understanding of the physical world, we incorporate questions that involve classical mechanics, electromagnetism, and other mainstream physical disciplines.
    \item Multiple discipline Visual Question Answering~\cite{yue2024mmmu, yue2025mmmupro}: We include complex questions spanning domains like science, art, and humanities within a visual context, to test the model's ability to leverage college-level subject knowledge and deliberate reason.
\end{itemize}

We visualize different tasks in the Multi-Modal Mixture Benchmark (M³-Bench) in Figure~\ref{fig:benchmark_overview}.

\subsubsection{Data Collection of Multi-Modal Mixture Benchmark}
\label{sec:Data Collection of Multi-Modal Mixture Benchmark}
To construct a diverse, high-quality evaluation benchmark aligned with our defined task taxonomy, we employ a dual-strategy methodology for dataset assembly. This methodology combines the reformatting of authoritative resources with targeted data collection. First, we reformat and integrate samples from established general-purpose and domain-specific benchmarks. This allows our benchmark to harness their inherent diversity and the validated expertise while reducing the requirement of instruction-following. Second, to address identified gaps, such as insufficient scale of visual concepts or specialized scenarios beyond the pre-trained model's capabilities, we collect new image-question pairs. This includes sourcing images of animals, plants, celebrities, and attractions from Wikipedia for visual concept identification, and harvesting web images to generate spatial reasoning tasks. The mixture methodology ensures that our benchmark is both comprehensive and precisely targeted.

\noindent \textbf{Reformatting Popular Benchmarks.} 
To encompass diverse perceptual and reasoning capabilities, we collect both general-purpose and specialized benchmarks of broad interest, including MMBench~\cite{liu2024mmbench}, MMMU-Pro~\cite{yue2025mmmupro}, MathVista~\cite{lu2024mathvista}, and Seephys~\cite{xiang2025seephys}.
These benchmarks usually require instruction-following abilities and are therefore unsuitable for the direct 0-shot evaluation of pretrained MLLMs.
As mentioned in Section~\ref{sec:Soft Discrimination Score}, we use SDS, which relies on images, questions, and a predefined set of candidates for 0-shot evaluation. So, we repurpose these benchmarks by unifying different question types into this standard format as follows:
\begin{itemize}
    \item For multiple-choice questions, considering the instability of pre-trained models in instruction following, we modify the prediction target from the option letter to the actual content of the option. This also mitigates biases from option order. Specifically, during evaluation, the model only receives images and questions as conditions and independently calculates the average token logits for each option. 
\item For free-form questions, we generate reasonable distractors. For numerical answers in MathVista, we apply random minor perturbations: ±1 for integers and ±0.1 for decimals. For expression-based answers in SeePhys, we use DeepSeek~\cite{liu2024deepseek} to generate distractors where only certain variables or quantities are altered while maintaining rationality.
    
\end{itemize}
To ensure a fair comparison of SDS scores across different tasks, we randomly sample distractors to make sure all questions maintain the same number of candidates. Specifically, 1 distractor per question in our implementation. This setup theoretically yields a random accuracy of 0.5. Empirically, we also find that LLaVA-OneVision-0.5B with a randomly initialized projector typically yields SDS scores around 0.5.
We discuss the impact of different distractor construction methods and questioning methods on results in the supplementary material.


\begin{figure}
\begin{centering}
\includegraphics[width=0.98\linewidth]{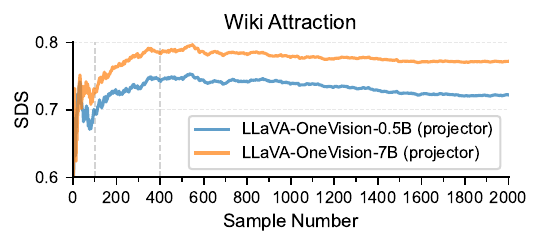}
\par\end{centering}
\vspace{-0.2cm}
\caption{\textbf{Impact of the number of test sample on model performance.} 
}
\label{fig:samplenum}
\end{figure}

\begin{figure*}[t]
\begin{centering}
\includegraphics[width=1.0\linewidth]{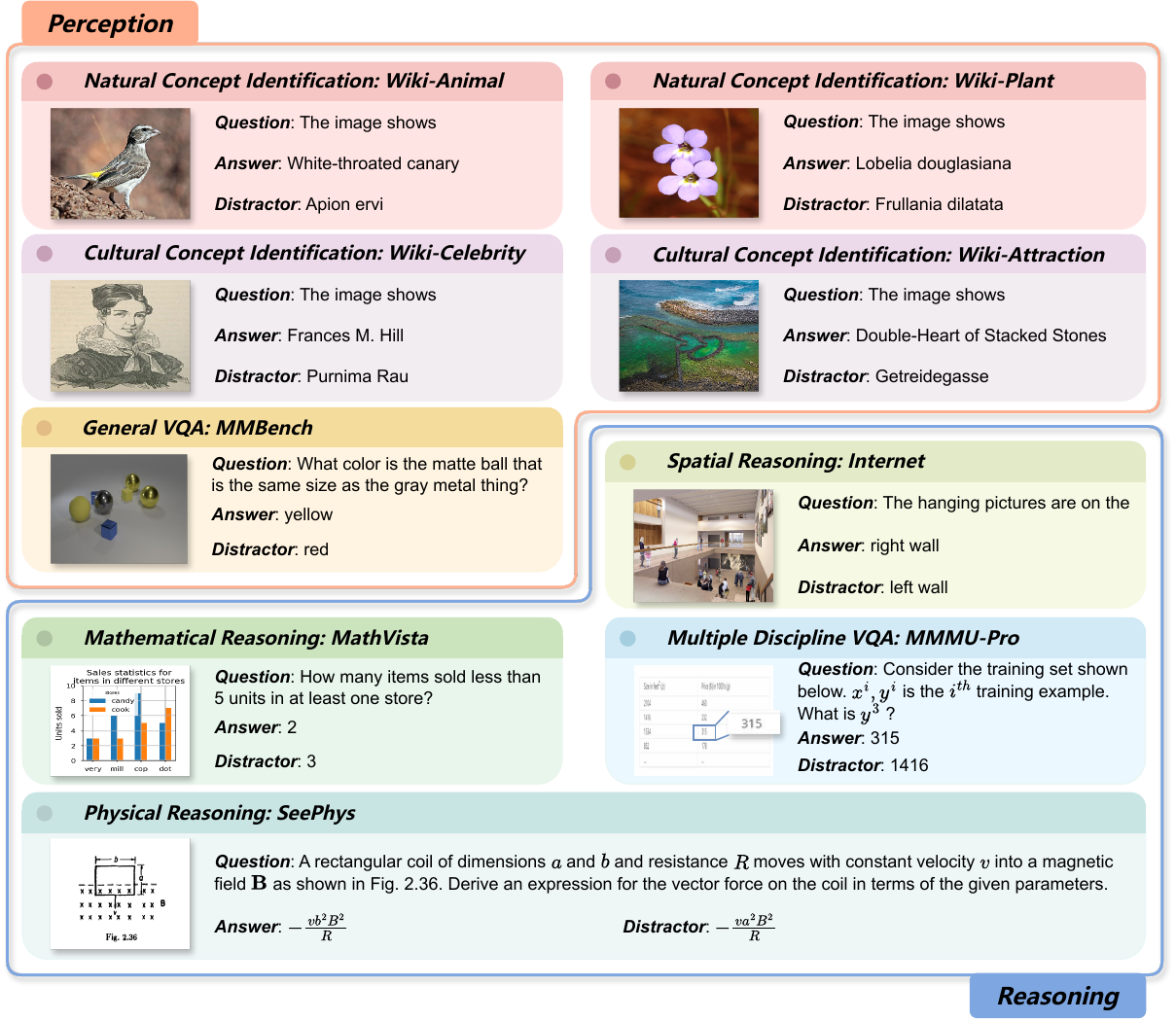}
\par\end{centering}
\vspace{-0.2cm}
\caption{\textbf{Digram of M³-Bench.} Our benchmark contains 7 task categories, collected from 9 sources. We sample QA pairs of each source for visualization. Each question is equipped with 1 answer and 1 distractor.}
\label{fig:benchmark_overview}
\end{figure*}

\begin{figure*}[t]
\begin{centering}
\includegraphics[width=1.0\linewidth]{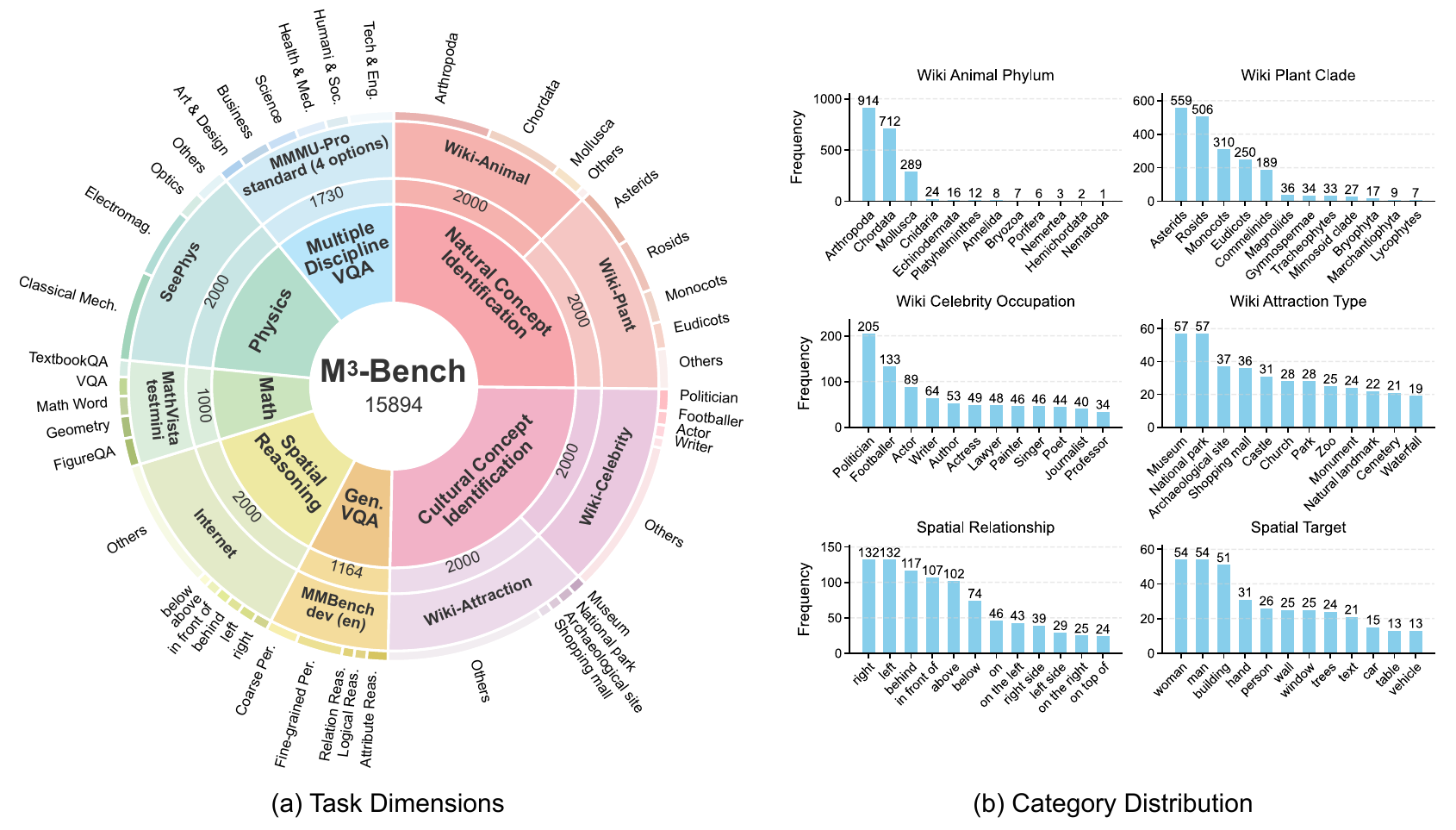}
\par\end{centering}
\vspace{-0.2cm}
\caption{\textbf{Data statistics of M³-Bench.} Section (a) presents the task dimensions, data sources, and problem count of M³-Bench. Section (b) shows the category distribution of our newly collected dataset (including the categories of animal, plant, celebrity, attraction collected from Wikipedia, and the categories of spatial relationship and spatial target collected from the web).
}
\label{fig:benchmark_statistics}
\end{figure*}

\noindent\textbf{Collecting new datasets.} We observe that the existing general benchmarks typically employ small sample sizes in terms of visual concepts, which may introduce significant random errors. Besides, they only require the model to recognize common visual concepts, which may lead to overly simplistic testing and fail to distinguish the strengths and weaknesses of different models (e.g., the ``Existence'' in MME~\cite{fu2025mme} and the ``Image Scene'' in MMBench~\cite{liu2024mmbench}). Given that Wikipedia is a highly reliable data source, we aim to evaluate pre-trained models' fine-grained understanding of broad visual concepts from Wikipedia to assess their perceptual capabilities. We consider the identification of natural and cultural concepts, and finally choose four categories: animal, plant, celebrity, and attraction.

Before formally constructing the dataset, we conduct a simple preliminary experiment to determine an appropriate sample size.
Specifically, we adopt a subset of 2,000 samples from the collected Wiki-Attraction category to test LLaVA-OneVision~\cite{li2024lavaonevision} with pre-trained projector. As shown in Figure~\ref{fig:samplenum}, we compute SDS scores across sample sizes ranging from 0 to 2,000. The performance curves of the two models intersect within the 0–100 sample range, indicating that evaluations with fewer than 100 samples are unreliable. Between 100 and 400 samples, performance differences exhibit considerable fluctuation, which could lead to unfair comparisons when evaluating multiple models. Beyond 400 samples, individual model performance variance decreases, with both the relative differences between models and their individual scores becoming consistent around 2,000 samples. Consequently, we adopt a sample size of 2,000 per category to ensure a reliable and diverse assessment.

To construct the dataset, we crawl images, descriptions, and titles of the four categories from Wikipedia. We design the question simply as ``The image shows'' to minimize the need for instruction-following, with the correct answer set to the Wikipedia title. Distractor options are randomly selected from other titles of the same category. We find that some retrieved images do not display the titles directly. For instance, instead of showing the plants themselves, they present maps of the plants' geographical distribution. We use \emph{Seed 1.6}\footnote{\url{https://seed.bytedance.com/en/seed1_6}} to determine whether the images contain titles, and for those without titles, we perform manual replacements.

During reformatting and utilizing domain-specific benchmarks, we observe that current spatial reasoning benchmarks are not ideal for evaluating multi-modal pre-training due to three primary issues:
\begin{itemize}
    \item Misalignment with the pre-training objective: Current benchmarks are designed as challenging downstream tasks with complex, realistic scenarios, where success typically requires specialized model architectures and training. Thus, they do not effectively assess foundational spatial reasoning capabilities.
    \item Data leakage: Benchmarks built upon classic datasets (e.g., COCO) risk significant data leakage between pre-training and evaluation phases, potentially leading to overfitting and inflated performance.
    \item Insufficient size: Small-scale benchmarks may pose risks of category biases and inaccurate comparisons.
\end{itemize}
Since the pre-training stage for MLLMs typically prioritizes broad data diversity over domain-specific depth, an effective spatial reasoning benchmark for this purpose must possess sufficient scale, diversity, and focus on fundamental skills. To achieve this goal, we implement the following pipeline: First, we crawl images from the internet to ensure diversity. Second, using a specialized MLLM to generate high-quality spatial descriptions for these images. Third, using these descriptions, we prompt DeepSeek~\cite{liu2024deepseek} to produce question-answer-distractor triplets concerning spatial reasoning, targeting one or more elements per image. Crucially, to eliminate ambiguity given the complexity of spatial relationships, we instruct the model to generate distractors that represent an opposite spatial relationship to the correct answer. Finally, we manually verify every item (image, question, answer, and distractor) for correctness.
The prompt for generating spatial reasoning tasks is provided in the supplementary material. Given the rapid expansion of MLLM pre-training data, complete avoidance of data leakage is practically infeasible. Nevertheless, we posit that evaluation outcomes retain robust comparative validity if they exhibit expected scaling trends between model sizes and avoid performance saturation (e.g., near-ceiling scores). Therefore, such results remain highly valuable for reference.

\subsubsection{Data Statistics of Multi-Modal Mixture Benchmark}
\label{Data Statistics of Multi-Modal Mixture Benchmark}

Figure~\ref{fig:benchmark_statistics} presents the data statistics of M³-Bench.

\noindent \textbf{Task Dimensions.}
In the present study, we have gathered data samples spanning across 7 task dimensions.
We depict the problem counts and data source of all tasks in Figure~\ref{fig:benchmark_statistics} (a). To ensure a reliable and comprehensive evaluation of each ability, we use at least 1k samples per data source. Specifically, for general VQA task, we use the dev english split from MMBench, which contains 1,164 questions. For mathematical reasoning, we use the testmini split from MathVista, with 1,000 questions. For physical reasoning, we use SeePhys, which contains 2,000 questions. For multiple discipline VQA, we use the standard (4 options) split of MMMU-Pro with 1,730 questions. For natural and cultural concept identification, we collect 2,000 samples per category from Wikipedia, yielding 8,000 questions. For spatial reasoning, we manually collect 2,000 samples from the internet. In total, our benchmark comprises 15,894 questions to fulfill a comprehensive evaluation on the perception and reasoning abilities of pre-trained MLLMs.

\noindent \textbf{Category Distribution.} 
We categorize concepts in our newly collected dataset using Wiki and DeepSeek~\cite{liu2024deepseek}, presenting the top 12 frequent categories in Figure~\ref{fig:benchmark_statistics} (b). The distribution of animal and plant categories aligns with real-world distributions: among animals, arthropods are most prevalent, while among plants, the asterid family shows significant dominance. While Wikipedia's taxonomy lists 18 animal and 21 plant categories, categories of celebrity and attraction extracted with DeepSeek are far more granular, reflecting their broader scope. Preliminary counts indicate 1,057 celebrity and 915 attraction categories.
For the spatial reasoning task, we analyze the two aspects: spatial relationship types and target description types. Spatial relationships include basic terms (e.g., \textit{right}, \textit{left}) and their common phrasal variants (e.g., \textit{on the left}, \textit{right side}), totaling 361 unique expressions in the answers. Target descriptions in questions and answers, primarily covering people and environments, comprise 2,382 distinct descriptions. We do not merge fine-grained categories (e.g., \textit{woman in a dress} and \textit{woman with a hat}), as our goal is to showcase data diversity rather than to impose a rigorous classification.


\begin{figure*}
\begin{centering}
\includegraphics[width=0.98\linewidth]{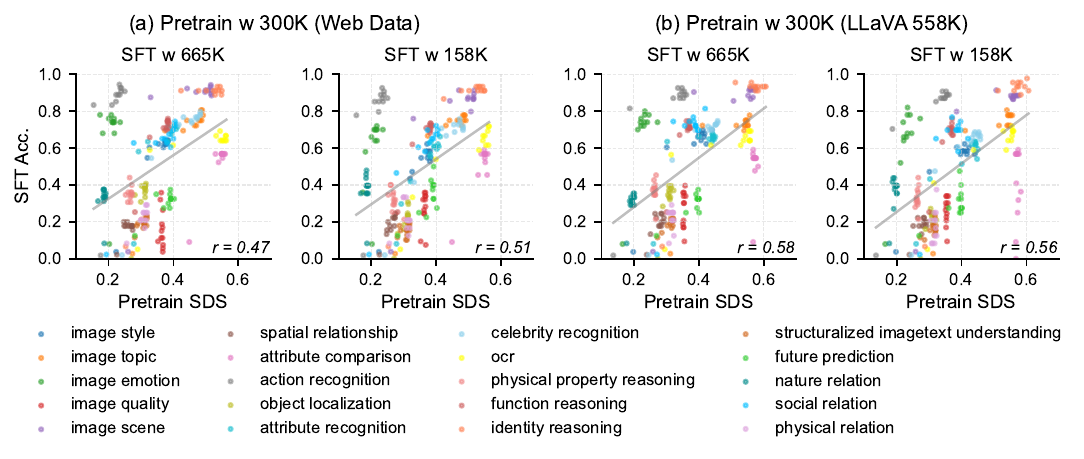}
\par\end{centering}
\vspace{-0.2cm}
\caption{\textbf{The positive correlation between pre-training score and finetuned performance.} We conduct the full MLLM training pipeline across different datasets. For pre-training, we utilize two general-domain data sources: 300K web data and 300K from LLaVA 558K. For fine-tuning, we use two data scales: LLaVA Instruct 665K and a randomly selected subset of 158K from 665K. During pre-training, we save a checkpoint every 100 steps and evaluate on 20 L3 category tasks in the reformatted MMBench-dev (en) split. Each checkpoint is subsequently fine-tuned and tested with the standard MMBench-dev (en) split. We plot scatter graphs with our introduced pre-training scores on the x-axis and fine-tuning scores on the y-axis, and calculate the Pearson correlation coefficient.}
\label{fig:pretrain_sft_scatter}
\end{figure*}

\section{Experiment}
\label{sec:Experiment}
In this section, we introduce the default training and evaluation settings (Sec.~\ref{sec:Experimental Setup}). We first conduct a series of pre-training experiments to explore whether the RADAR pre-training evaluation can reflect the performance after fine-tuning. (Sec.~\ref{sec:Effectiveness of RADAR}). Then we evaluate open-source pretrained MLLMs with RADAR and investigate the impact of model parameter, data scale, and pre-training strategy, revealing a key finding of asymmetric ability development (Sec.~\ref{sec:Differential Scaling Law}). With the guidance of this finding, we further conduct controlled two-stage pre-training experiments to 
explore how abilities react with adding homogeneous data (Sec.~\ref{sec:Differential Data Learning Trajectories}).

\subsection{Experimental Setup}
\label{sec:Experimental Setup}
This section details our default training, evaluation, and dataset configurations. 

\noindent\textbf{Multi-modal Pre-training Setup.} For pre-training, we follow the standard LLaVA setting: only the MLP projector is trained while the rest of the model is frozen, with the learning rate of 1e-3 and the global batch size of 256. We use Qwen2-0.5B-Instruct~\cite{yang2024qwen2} as the base LLM and the SigLIP-so400m-patch14-384~\cite{zhai2023siglip} as the vision encoder.

\noindent\textbf{Multi-modal Pre-training data.} Our pre-training corpus comprises general and specialized data:
\begin{itemize}
\item General Domain: (1) 300K web-crawled image-text pairs across diverse categories, and (2) 300K samples randomly selected from LLaVA Pre-train LCS-558K dataset.
\item Wikipedia Concepts: For the plant and celebrity categories, we use the corresponding Wikipedia description as the pre-training label.
\item Spatial Reasoning: We use the spatial description generated by our specialized MLLM as the pre-training label.
\item Mathematics: We use the math subset from R1-Onevision~\cite{yang2025r1onevision}, training only on the answer text.
\end{itemize}
For the specialized Wiki and spatial reasoning datasets, we hold out 2K samples for benchmark construction and sample 50K for training, ensuring no overlap.

\noindent\textbf{Fine-tuning Setup.} For supervised fine-tuning (SFT), we employ the LLaVA SFT 665K dataset with the learning rate of 2e-5 and the global batch size of 128.

\noindent All experiments are conducted with V100 GPUs and the random seed is set to 0.

\noindent\textbf{Open-source Models for Evaluation.} Our evaluation experiments involve three series of open-source pre-trained MLLMs: LLaVA-Onevision (projectors) 0.5B and 7B, Qwen2-VL 2B and 7B, and InternVL-3.5-Pretrained spanning 5 model scales: 1B, 2B, 4B, 8B, 14B. During evaluation, LLaVA-OneVision (projectors) encodes each image to 729 tokens. For InternVL-3.5 and Qwen2-VL, we set the maximum number of tokens per image to 768. To maintain consistency with the pre-training format, we adapt the evaluation input using the same conversation templates as the open-source models.
Details on their model size and pre-training data volume are provided in the supplementary material.

\noindent\textbf{Evaluation Setup.} In Section~\ref{sec:Effectiveness of RADAR}, we use the reformatted MMBench-dev (en) split~\cite{liu2024mmbench} with its original number of distractors, following its L3-level ability taxonomy for analysis. In Sec~\ref{sec:Revealing Asymmetric Ability Development with RADAR}, we use the proposed M³-Bench, which limits the number of distractors for each question to 1 for fair comparison with different abilities. For standard evaluation with finetuned models, we use VLMEvalKit~\cite{duan2024vlmevalkit}.

\begin{table*}[t]
\centering
\caption{Performance comparison results on perception abilities of M³-Bench. The best results are annotated in \textbf{bold} fonts.}
\renewcommand\arraystretch{1}
\resizebox{0.98\textwidth}{!}{
\setlength{\tabcolsep}{3.3mm}{
\begin{tabular}{l|cc|cc|cc|cc|cc|cc} \toprule
\multirow{2}{*}{Method}& \multicolumn{2}{c|}{Average} & \multicolumn{2}{c|}{MMBench} & \multicolumn{2}{c|}{Wiki-Animal} & \multicolumn{2}{c|}{Wiki-Plant} & \multicolumn{2}{c|}{Wiki-Celebrity} & \multicolumn{2}{c}{Wiki-Attraction} \\ & Acc. & \textbf{SDS} & Acc. & \textbf{SDS} & Acc. & \textbf{SDS} & Acc. & \textbf{SDS} & Acc. & \textbf{SDS} & Acc. & \textbf{SDS} \\ 
\midrule
LLaVA-OneVision-projector-0.5B \cite{li2024lavaonevision} & 0.627 & 0.608 & 0.664 & 0.620 & 0.560 & 0.551 & 0.541 & 0.540 & 0.626 & 0.606 & 0.744 & 0.722\\
LLaVA-OneVision-projector-7B \cite{li2024lavaonevision} & 0.703 & 0.671 & 0.713 & 0.674 & 0.669 & 0.634 & 0.636 & 0.610 & 0.692 & 0.665 & 0.806 & 0.772 \\
Intern3.5-VL-Pretrained-1B \cite{wang2025internvl3_5} & 0.626 & 0.606 & 0.711 & 0.660 & 0.579 & 0.565 & 0.546 & 0.548 & 0.603 & 0.585 & 0.691 & 0.668 \\
Intern3.5-VL-Pretrained-2B \cite{wang2025internvl3_5} & 0.671 & 0.646 & 0.758 & 0.693 & 0.639 & 0.621 & 0.602 & 0.594 & 0.628 & 0.615 & 0.730 & 0.706 \\
Intern3.5-VL-Pretrained-4B \cite{wang2025internvl3_5} & 0.700 & 0.668 & 0.766 & 0.707 & 0.701 & 0.668 & 0.640 & 0.617 & 0.663 & 0.636 & 0.730 & 0.712 \\
Intern3.5-VL-Pretrained-8B \cite{wang2025internvl3_5} & 0.723 & 0.694 & 0.789 & 0.726 & 0.704 & 0.681 & 0.654 & 0.636 & 0.693 & 0.677 & 0.775 & 0.751 \\
Intern3.5-VL-Pretrained-14B \cite{wang2025internvl3_5} & 0.736 & 0.709 & 0.789 & 0.727 & 0.716 & 0.698 & 0.672 & 0.658 & 0.715 & 0.691 & 0.789 & 0.768  \\
Qwen2-VL-2B \cite{wang2024qwen2vl} & 0.707 & 0.694 & 0.716 & 0.681 & 0.650 & 0.648 & 0.654 & 0.644 & 0.681 & 0.679 & 0.834 & 0.819 \\  
Qwen2-VL-7B \cite{wang2024qwen2vl} & 0.758 & \textbf{0.736} & 0.790 & \textbf{0.735} & 0.734 & \textbf{0.717} & 0.707 & \textbf{0.687} & 0.715 & \textbf{0.709} & 0.846 & \textbf{0.830} \\
\bottomrule
\end{tabular}}
}

\label{table:perc_opensource_res}
\vspace{-0.2cm}
\end{table*}

\begin{table*}[t]
\centering
\caption{Performance comparison results on reasoning abilities of M³-Bench. The best results are annotated in \textbf{bold} fonts.}
\renewcommand\arraystretch{1}
\resizebox{0.98\textwidth}{!}{
\setlength{\tabcolsep}{3.3mm}{
\begin{tabular}{l|cc|cc|cc|cc|cc} \toprule
\multirow{2}{*}{Method}& \multicolumn{2}{c|}{Average} & \multicolumn{2}{c|}{Spatial Reas.} & \multicolumn{2}{c|}{MathVista} & \multicolumn{2}{c|}{SeePhys} & \multicolumn{2}{c}{MMMU-Pro} \\ & Acc. & \textbf{SDS} & Acc. & \textbf{SDS} & Acc. & \textbf{SDS} & Acc. & \textbf{SDS} & Acc. & \textbf{SDS} \\ 
\midrule
LLaVA-OneVision-projector-0.5B \cite{li2024lavaonevision} & 0.506 & 0.508 & 0.526 & 0.514 & 0.521 & 0.509 & 0.453 & 0.488 & 0.526 & 0.520 \\
LLaVA-OneVision-projector-7B \cite{li2024lavaonevision} & 0.514 & 0.515 & 0.549 & 0.531 & 0.525 & 0.513 & 0.446 & 0.486 & 0.536 & 0.530 \\
Intern3.5-VL-Pretrained-1B \cite{wang2025internvl3_5} & 0.571 & 0.535 & 0.624 & 0.554 & 0.613 & 0.570 & 0.492 & 0.491 & 0.557 & 0.527 \\
Intern3.5-VL-Pretrained-2B \cite{wang2025internvl3_5} & 0.603 & 0.561 & 0.667 & 0.589 & 0.680 & 0.621 & 0.503 & 0.497 & 0.562 & 0.536 \\
Intern3.5-VL-Pretrained-4B \cite{wang2025internvl3_5} & 0.647 & 0.589 & 0.756 & 0.665 & 0.703 & 0.641 & 0.535 & 0.498 & 0.594 & 0.552 \\
Intern3.5-VL-Pretrained-8B \cite{wang2025internvl3_5} & 0.670 & 0.609 & 0.767 & 0.685 & 0.767 & 0.691 & 0.536 & 0.504 & 0.611 & 0.554 \\
Intern3.5-VL-Pretrained-14B \cite{wang2025internvl3_5} & 0.676 & \textbf{0.614} & 0.791 & \textbf{0.704} & 0.731 & 0.665 & 0.559 & \textbf{0.510} & 0.625 & \textbf{0.578} \\
Qwen2-VL-2B \cite{wang2024qwen2vl} & 0.599 & 0.561 & 0.655 & 0.575 & 0.688 & 0.633 & 0.489 & 0.501 & 0.563 & 0.537 \\  
Qwen2-VL-7B \cite{wang2024qwen2vl} & 0.656 & 0.606 & 0.752 & 0.663 & 0.768 & \textbf{0.700} & 0.527 & 0.505 & 0.576 & 0.556 \\
\bottomrule
\end{tabular}}
}

\label{table:reason_opensource_res}
\vspace{-0.2cm}
\end{table*}

\subsection{Does the RADAR pre-training evaluation reflect the downstream performance?}
\label{sec:Effectiveness of RADAR}
The pre-training phase serves as the critical foundation for post-training. Consequently, an effective evaluation of pre-training quality should be indicative of final model performance after fine-tuning. To investigate whether our proposed RADAR evaluation framework reflects downstream performance, we conduct the full MLLM training pipeline across different datasets. For pre-training, we utilize two general-domain data sources: 300K web-crawled and 300K from LLaVA-558K. For fine-tuning, we use two data scales: LLaVA Instruct 665K and a randomly selected subset of 158K from 665K to examine different ratios of pre-training and fine-tuning data. During pre-training, we save a checkpoint every 100 steps. Each checkpoint is subsequently fine-tuned. This procedure allows us to evaluate the model with two strategies: first, assessing the pre-trained model via RADAR on the reformatted MMBench, and second, measuring the fine-tuned model’s performance via standard evaluation on the original MMBench.

Figure~\ref{fig:pretrain_sft_scatter} presents scatter plots comparing these pre-training and finetuning scores for each L3 ability in MMBench. The 4 experimental configurations (2 pre-training datasets × 2 fine-tuning datasets) yield 11 checkpoints each, evaluated across 20 L3 capabilities, for a total of 880 data points.

\textbf{Positive correlation between pre-training and finetuning performance.} The Pearson correlation coefficients computed in all four groups reveal a positive correlation between pre-training and fine-tuning performance. This demonstrates that RADAR reliably reflects pre-training quality and is generally predictive of final model performance across varying data configurations.
We also note that for certain capabilities, pre-training scores are weak predictors of fine-tuning performance. We primarily attribute this to a distributional shift between pre-training and fine-tuning data, which can lead to specific abilities developing predominantly in one phase. For example, \textit{image quality} in each group remains static near 0.38 during pre-training but improves from ~0.1 to ~0.4 after fine-tuning.

\subsection{Revealing Asymmetric Ability Development with RADAR}
\label{sec:Revealing Asymmetric Ability Development with RADAR}

\subsubsection{How do data and parameter scale impact different ability development?}
\label{sec:Differential Scaling Law}
Having verified RADAR's effectiveness, we evaluate three series of open-source pre-trained MLLMs and analyze the impact of data and model size on different abilities.

\textbf{Open-source MLLM evaluation.} Table~\ref{table:perc_opensource_res} and Table~\ref{table:reason_opensource_res} present the performance of open-source models on the perception and reasoning tasks of M³-Bench, respectively.
By comparing accuracy and SDS, we observe that the two metrics yield comparable average model rankings. However, accuracy is usually higher than SDS, indicating that it fails to distinguish between fortuitous and robust success, thereby overestimating the model's true capability.

Table~\ref{table:perc_opensource_res} and Table~\ref{table:reason_opensource_res} also reveal a notable divergence in performance between perception and reasoning tasks. On perception tasks (Table~\ref{table:perc_opensource_res}), Qwen2-VL-7B, trained on the largest pre-training dataset ($\sim$1.4T tokens), achieves the best performance in each task. On average, it even outperforms the larger InternVL-3.5-Pretrained-14B by approximately 3 points. Conversely, on reasoning tasks (Table~\ref{table:reason_opensource_res}), InternVL-3.5-Pretrained-14B, the largest model with the latest generation of language model (Qwen3~\cite{yang2025qwen3}), excels in most tasks and attains the highest overall average score.

This divergence raises a key question: how do model scale and data scale influence the performance of open-source models in perception and reasoning tasks?
To investigate this relationship, we plot the models' average performance on perception and reasoning tasks against data volume and model size, as shown in Figure~\ref{fig:scaling_law_perc-vs-reas}.

\textbf{Impact of model size.} Increasing the model parameters has an overall enhancing effect on perception and reasoning. However, the trainable parameters seem to be more crucial for enhancing reasoning. Increasing trainable parameters has a nearly identical promoting effect on perception and reasoning, showing almost parallel fitted curves.

\textbf{Impact of data volume.} In contrast, increasing pre-training data yields different effects for perception and reasoning. For perception, increasing data size generally improves model performance across various model designs and sizes. For reasoning, however, scaling data does not enable the Qwen2-VL series to surpass the InternVL-3.5 series, as it does in perception. This plateau suggests that the primary bottleneck in reasoning performance may lie beyond data volume.

\textbf{Impact of pre-training strategies.} Moving to pre-training strategies, we compare LLaVA-OneVision, which employs lightweight alignment training on the projector layer, with the fully open, large-scale pre-training of the Qwen2-VL and InternVL-3.5 series. Results indicate that lightweight alignment training on limited data yields notable gains in perception tasks but offers almost no improvement in reasoning. In contrast, large-scale pre-training substantially improves performance across both perception and reasoning. These findings underscore the importance of fully open, large-scale pre-training for developing the comprehensive capabilities of vision-language models.

\textbf{Discussion.} Based on the experimental results, we reveal several key insights that can guide future work in MLLM pre-training:
First, the influence of model and data scale is not uniform across different capabilities, suggesting that future scaling analyses should be ability-decomposed and that a singular scaling law is insufficient. The primary bottleneck for advanced reasoning is likely to shift from data volume to model architecture, optimization, or data quality.
Second, lightweight alignment on limited, curated data is a suboptimal foundation. While effective for rapid adaptation in perception-dominated tasks, it fails to engender robust reasoning capabilities. Open, large-scale pre-training on diverse data is fundamental for developing comprehensive and balanced model abilities.
Consequently, future MLLM pre-training strategy should prioritize fully open, large-scale foundational training, as well as capability-aware scaling that optimizes model capacity and data composition for targeted abilities.

\begin{figure*}
\begin{centering}
\includegraphics[width=0.98\linewidth]{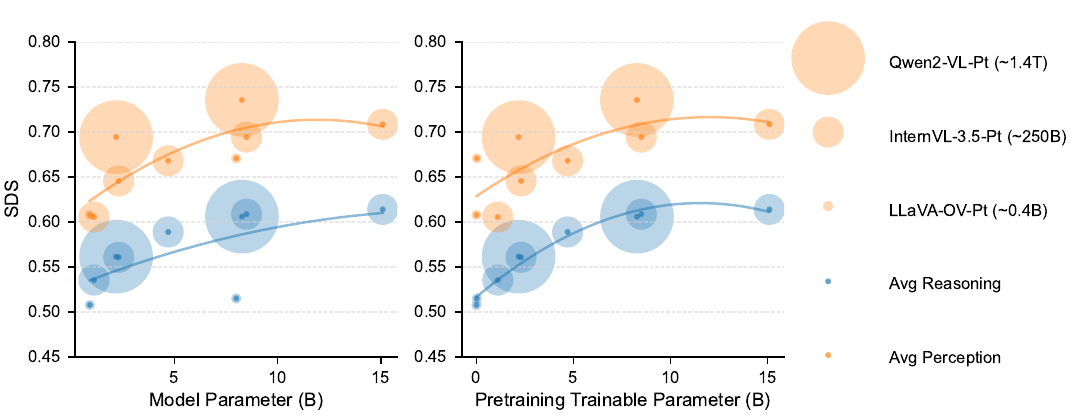}
\par\end{centering}
\vspace{-0.2cm}
\caption{\textbf{Impact of model size, data volume, and pretraining strategies.} The plots illustrate the average perception and reasoning performance of open-source MLLMs, as a function of the amount of model parameters and pre-training trainable parameters (model size). The circle's area represents the amount of pre-training data (data size). We use the LLaVA-OneVision series as a representative of lightweight alignment pre-training on the projector layer, and the Qwen2-VL and InternVL-3.5 series as representatives of full open, large-scale pre-training. ``Pt'' denotes their pretrained model.
The trend shows asymmetric improvement of perception and reasoning with respect to model size, data volume, and pretraining strategies.}
\label{fig:scaling_law_perc-vs-reas}
\end{figure*}

\begin{figure*}[t]
\begin{centering}
\includegraphics[width=0.98\linewidth]{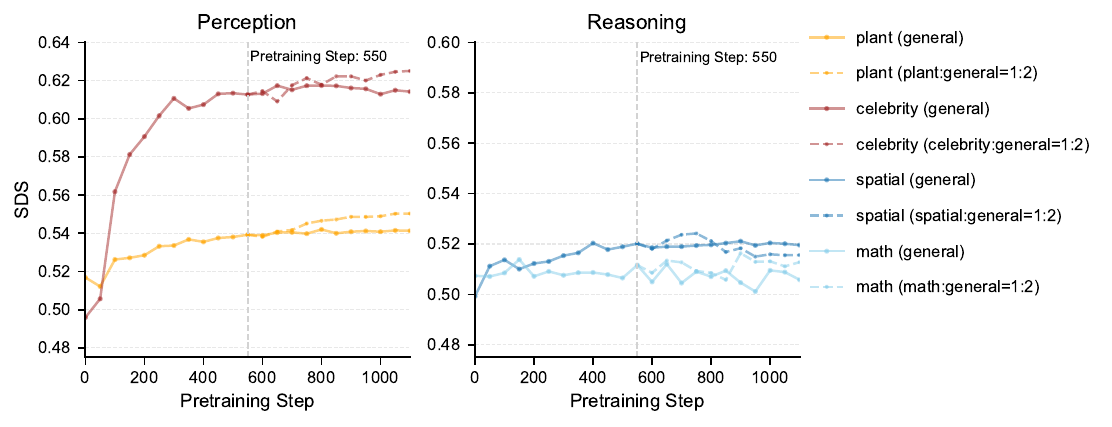}
\par\end{centering}
\vspace{-0.2cm}
\caption{\textbf{Impact of increasing data of the corresponding category on the ability.} First, we conduct a full training using 300K general data. Then, we use the weights at the middle position (550 steps) as the initial weights. We adjust the data to a mixed single-category of 50K and a general of 100K, and carry out the second stage of training (presented by the dotted line). In the legend, the colors of the curves represent the evaluation task category, and the information in parentheses indicates the pre-training data category.}
\label{fig:two_stage_perc-vs-reas}
\end{figure*}

\subsubsection{How does adding homogeneous data affect the improvements in the model's capabilities during subsequent training?}
\label{sec:Differential Data Learning Trajectories}

Based on our earlier analysis, the LLaVA-style pre-training is inadequate for improving reasoning performance. To investigate whether this limitation stems from the perception-heavy composition of the LLaVA-558K dataset, we design a controlled two-stage pre-training experiment. In this experiment, we systematically introduce data from specific capability categories in the second stage and track the resulting changes in model performance.
Specifically, we use Wiki-Plant and Wiki-Celebrity identification as proxy tasks for perception, and spatial and mathematical reasoning as proxy tasks for reasoning. The pre-training consists of two stages: the first stage (0–550 steps) uses LLaVA-558K as general pre-training data. In the second stage (550–1,100 steps), we set the mixture ratio of task-specific to general data to 1:2 to increase the proportion of targeted data during subsequent pre-training.

\textbf{Comparison between increasing perception and reasoning data}. Figure~\ref{fig:two_stage_perc-vs-reas} presents the pre-training performance curves. The results show that perception and reasoning tasks respond differently to targeted data augmentation. For perception tasks, specifically plant and celebrity identification, adding category-specific data in the second stage reliably improves performance. In contrast, reasoning tasks exhibit significant fluctuations under the same conditions: mathematical reasoning shows modest and inconsistent gains, while spatial reasoning shows only fluctuations.
This controlled experiment demonstrates that with identical models and training methods, perception and reasoning exhibit markedly different learning trajectories as homogeneous data increases. Compared to the more straightforward methods effective for improving perception, enhancing reasoning ability appears to require a multifaceted optimization of factors, including model scale, data volume, architectural design (e.g., the choice of backbone language model), and training strategy.

\textbf{Comparison between answer ranking accuracy and SDS}. Controlled experiments provide an ideal scenario for validating metrics on tracking the acquisition of model capabilities. To compare with our proposed RADAR method, we also plot the performance curves of the two-stage pre-training using logits-based answer-ranking accuracy, as shown in Figure~\ref{fig:two_stage_perc-vs-reas_acc}.
We find that accuracy is an insufficient metric for tracking changes in model capability. For perception tasks, accuracy fails to capture the model's increasing preference for correct answers: it shows minor improvement in celebrity identification and noisy fluctuations in plant identification. For reasoning tasks, accuracy is even more unstable, hindering clear analysis. In contrast, RADAR captures nuanced shifts in model preference while maintaining stability, making it better suited for tracking the acquisition of different capabilities throughout the pre-training process.

\begin{figure*}
\begin{centering}
\includegraphics[width=0.98\linewidth]{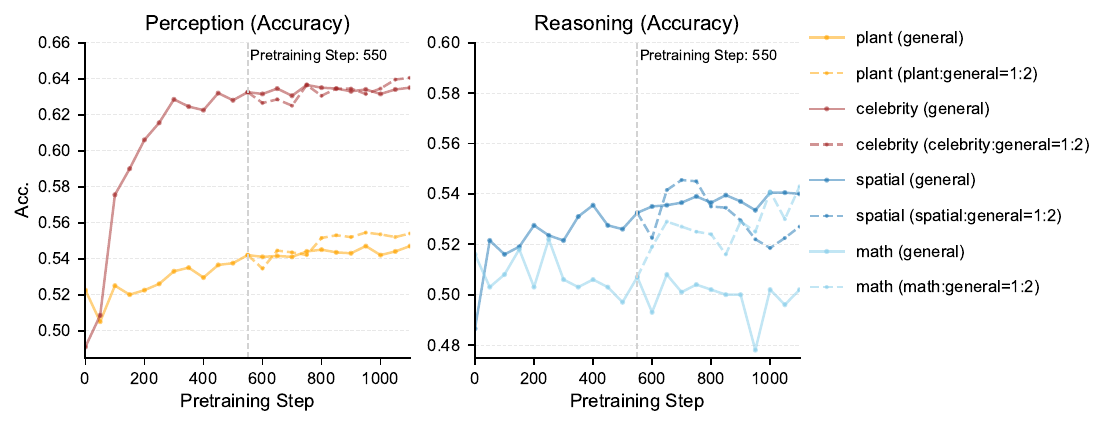}
\par\end{centering}
\vspace{-0.2cm}
\caption{\textbf{Accuracy as the indicator of ability evolution.} We evaluate each intermediate checkpoint with logits-based answer ranking strategy to compute the accuracy.
}
\label{fig:two_stage_perc-vs-reas_acc}
\end{figure*}

\section{Conclusion}

In this work, we propose \textbf{RADAR}, an efficient ability-centric evaluation framework for revealing asymmetric ability development for MLLM pre-training.
By introducing the soft discrimination score and the multi-modal mixture benchmark, RADAR provides robust, discriminative measurements of multi-dimensional pre-training performance that effectively reflect fine-tuning performance.
Guided by extensive assessments, we reveal the phenomenon of the asymmetry in the development of perception and reasoning abilities across diverse factors, including data volume, model size, and pretraining strategy.
The controlled two-stage experimental results further highlight the heterogeneity of capability learning in the MLLM pre-training process. We believe that our RADAR can provide meaningful references for designing optimal MLLM pre-training strategies to benefit future multi-modal base model research.

\bibliography{main}

\begin{thebibliography}{1}
\bibliographystyle{IEEEtran}

\bibitem{ref1}
{\it{Mathematics Into Type}}. American Mathematical Society. [Online]. Available: https://www.ams.org/arc/styleguide/mit-2.pdf

\bibitem{ref2}
T. W. Chaundy, P. R. Barrett and C. Batey, {\it{The Printing of Mathematics}}. London, U.K., Oxford Univ. Press, 1954.

\bibitem{ref3}
F. Mittelbach and M. Goossens, {\it{The \LaTeX Companion}}, 2nd ed. Boston, MA, USA: Pearson, 2004.

\bibitem{ref4}
G. Gr\"atzer, {\it{More Math Into LaTeX}}, New York, NY, USA: Springer, 2007.

\bibitem{ref5}M. Letourneau and J. W. Sharp, {\it{AMS-StyleGuide-online.pdf,}} American Mathematical Society, Providence, RI, USA, [Online]. Available: http://www.ams.org/arc/styleguide/index.html

\bibitem{ref6}
H. Sira-Ramirez, ``On the sliding mode control of nonlinear systems,'' \textit{Syst. Control Lett.}, vol. 19, pp. 303--312, 1992.

\bibitem{ref7}
A. Levant, ``Exact differentiation of signals with unbounded higher derivatives,''  in \textit{Proc. 45th IEEE Conf. Decis.
Control}, San Diego, CA, USA, 2006, pp. 5585--5590. DOI: 10.1109/CDC.2006.377165.

\bibitem{ref8}
M. Fliess, C. Join, and H. Sira-Ramirez, ``Non-linear estimation is easy,'' \textit{Int. J. Model., Ident. Control}, vol. 4, no. 1, pp. 12--27, 2008.

\bibitem{ref9}
R. Ortega, A. Astolfi, G. Bastin, and H. Rodriguez, ``Stabilization of food-chain systems using a port-controlled Hamiltonian description,'' in \textit{Proc. Amer. Control Conf.}, Chicago, IL, USA,
2000, pp. 2245--2249.

\end{thebibliography}

\bibliographystyle{IEEEtran}
\bibdata{IEEEabrv,egbib}	

\begin{IEEEbiography}[{\includegraphics[width=1in,height=1.25in,clip,keepaspectratio]{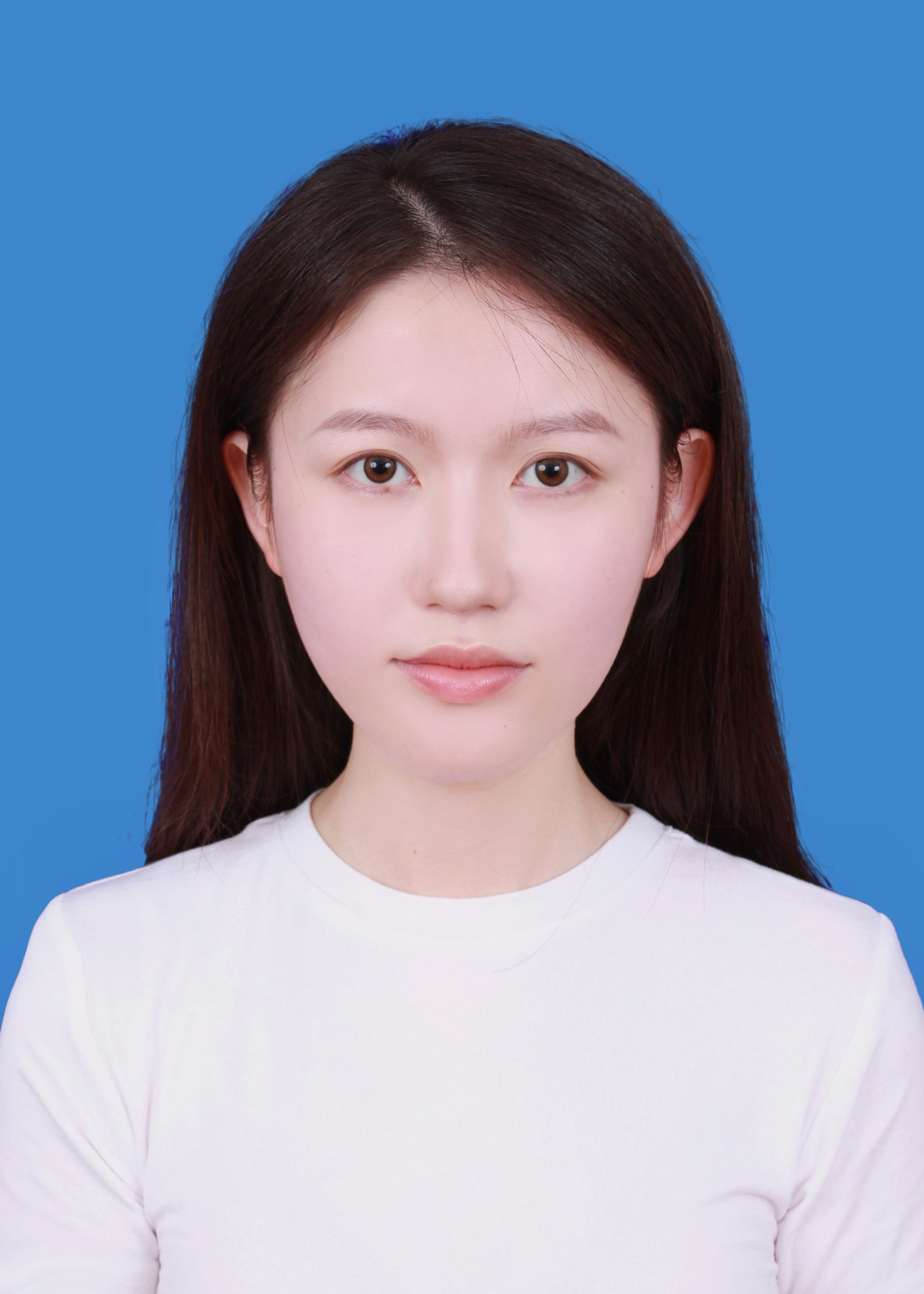}}]{Yunshuang Nie} received the B.E. degree in Sun Yat-sen University, Shenzhen, China, in 2023. She is currently working toward the M.E. in the school of intelligent systems engineering of Sun Yat-sen University. Her current research interests include vision-and-language understanding and embodied AI.
\end{IEEEbiography}

\begin{IEEEbiography}[{\includegraphics[width=1in,height=1.25in,clip,keepaspectratio]{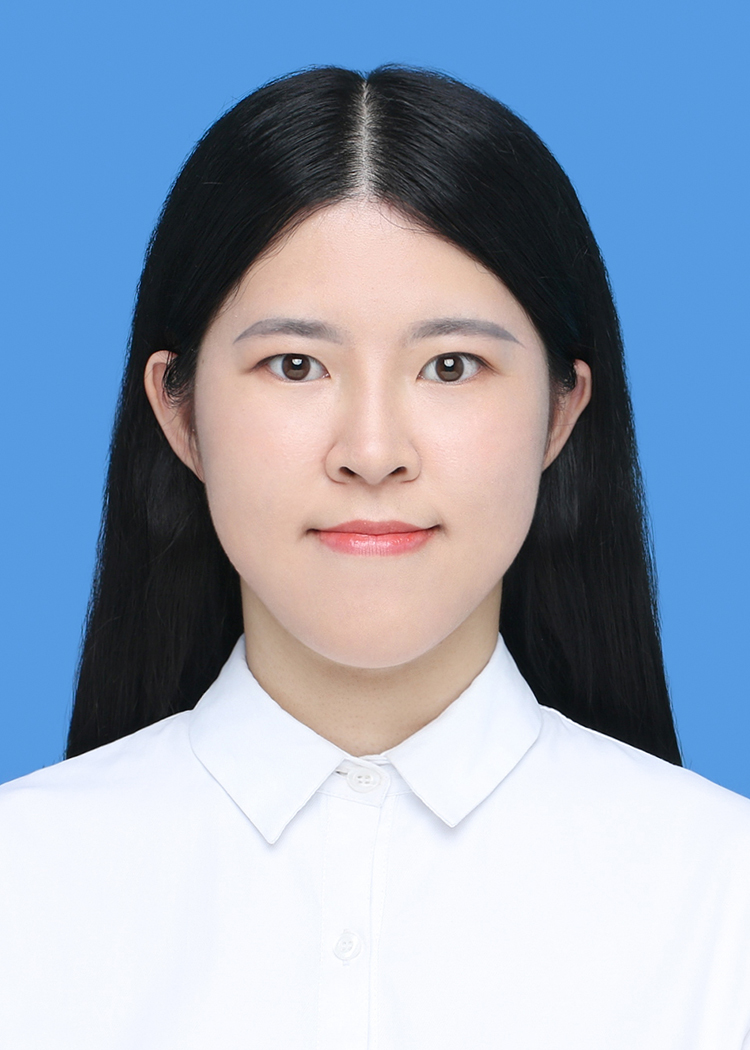}}]
{Bingqian Lin} is currently a postdoc researcher at Shanghai Jiao Tong University, advised by Prof. Cewu Lu. She received her PhD degree from Sun Yat-sen University in 2024, advised by Prof. Xiaodan Liang and Prof. Liang Lin. She received the B.E. and the M.E.
degree in Computer Science from University of
Electronic Science and Technology of China and
Xiamen University, in 2016 and 2019, respectively.
 Her research interests include vision-and-language understanding and embodied AI.
\end{IEEEbiography}

\begin{IEEEbiography}
[{\includegraphics[width=1in,height=1.25in, clip,keepaspectratio]{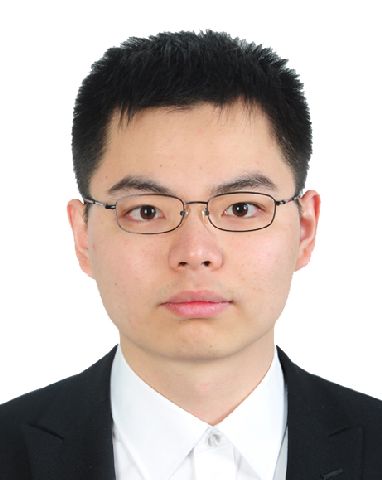}}]{Minzhe Niu} is currently a researcher with Yinwang Intelligent Technology Co., Ltd. He received his B.E. and M.E. degrees in Shanghai Jiao Tong University. His research interest includes multi-modality learning, autonomous driving and machine learning.
\end{IEEEbiography}

\begin{IEEEbiography}
[{\includegraphics[width=1in,height=1.25in, clip,keepaspectratio]{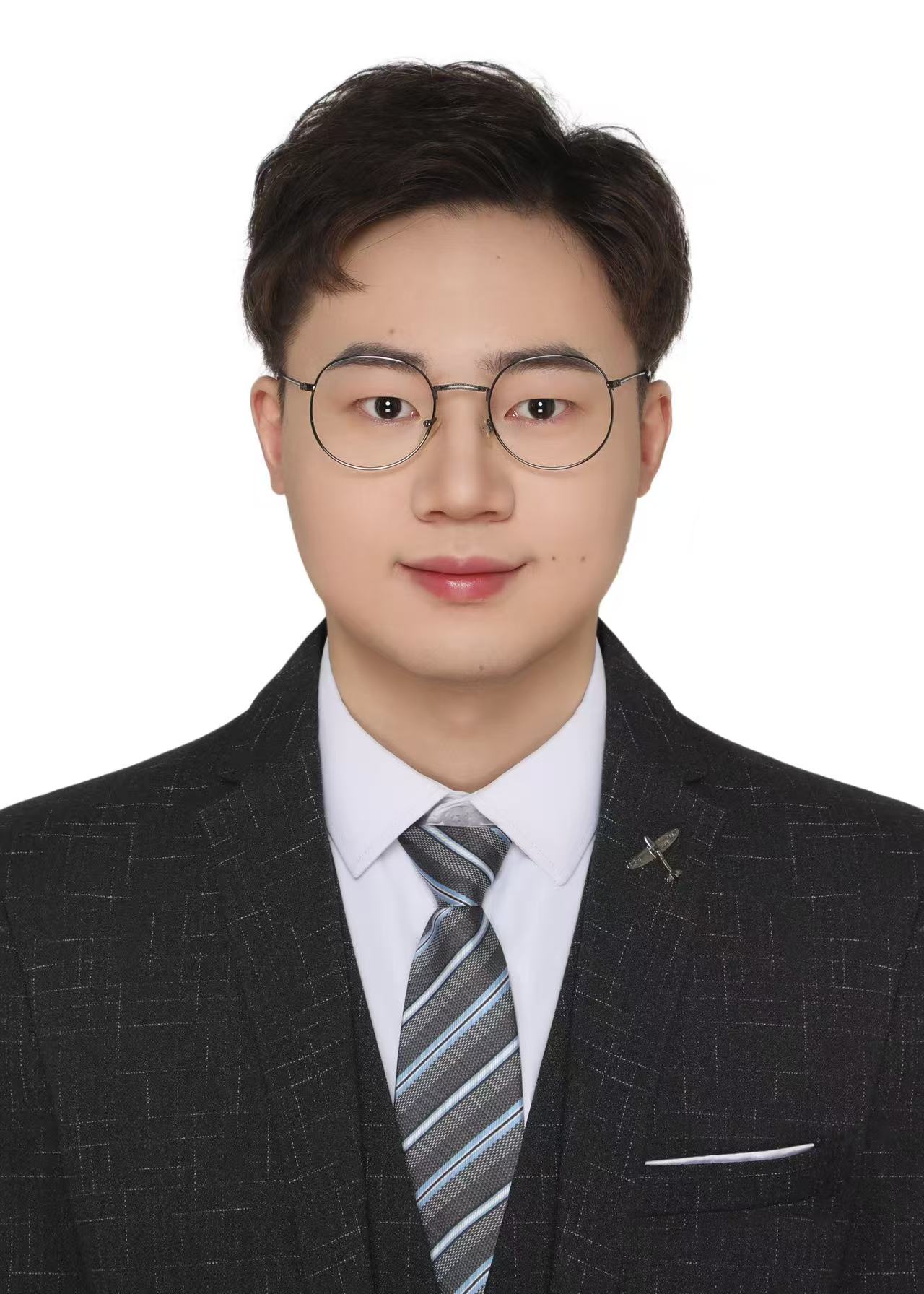}}]{Kun Xiang} is currently a PhD student at HCP-I2
Lab in Sun Yat-sen University advised by Prof.
Xiaodan Liang. He has received his B.S. degree and M.S. degree from School of Intelligent
Systems Engineering in Sun Yat-sen University,
China, in 2021 and 2024, respectively. He is interested in generalizable multimodal AI systems
and high order reasoning capability in MLLMs.
\end{IEEEbiography}

\begin{IEEEbiography}
[{\includegraphics[width=1in,height=1.25in, clip,keepaspectratio]{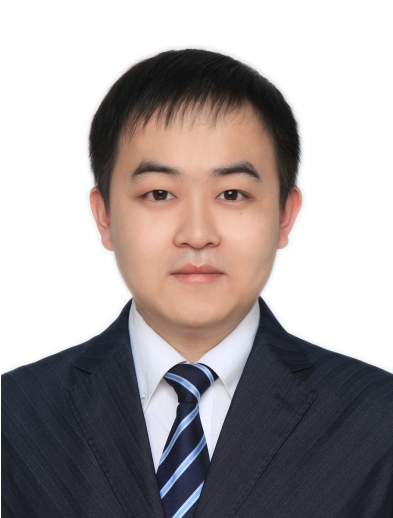}}]{Jianhua Han} received the Bachelor Degree in 2016 and Master Degree in 2019 from Shanghai Jiao Tong University, China. He is currently a researcher with Yinwang Intelligent Technology Co., Ltd. His research interests lie primarily in deep learning and computer vision.
\end{IEEEbiography}

\begin{IEEEbiography}
[{\includegraphics[width=1in,height=1.25in, clip,keepaspectratio]{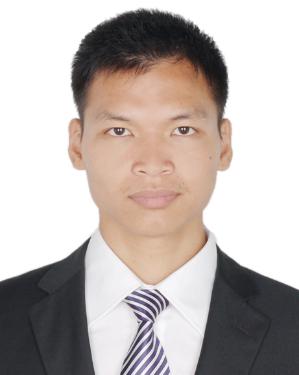}}]{Guowei Huang} is currently a researcher at Huawei's 2012 Lab. He received his M.E. degree from Xiamen University. His research interests include embodied intelligence, whole-body control, and parallel computing.
\end{IEEEbiography}

\begin{IEEEbiography}
[{\includegraphics[width=1in,height=1.25in, clip,keepaspectratio]{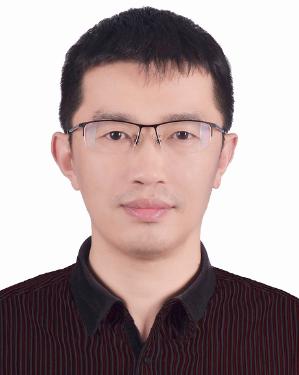}}]{Xingyue Quan} is currently a Senior Program Director for Embodied AI at Huawei’s 2012 Laboratories. Since 2017, he has been actively engaged in AI application research and innovation, and currently leads the Embodied AI research and innovation initiatives.
\end{IEEEbiography}

\begin{IEEEbiography}
[{\includegraphics[width=1in,height=1.25in, clip,keepaspectratio]{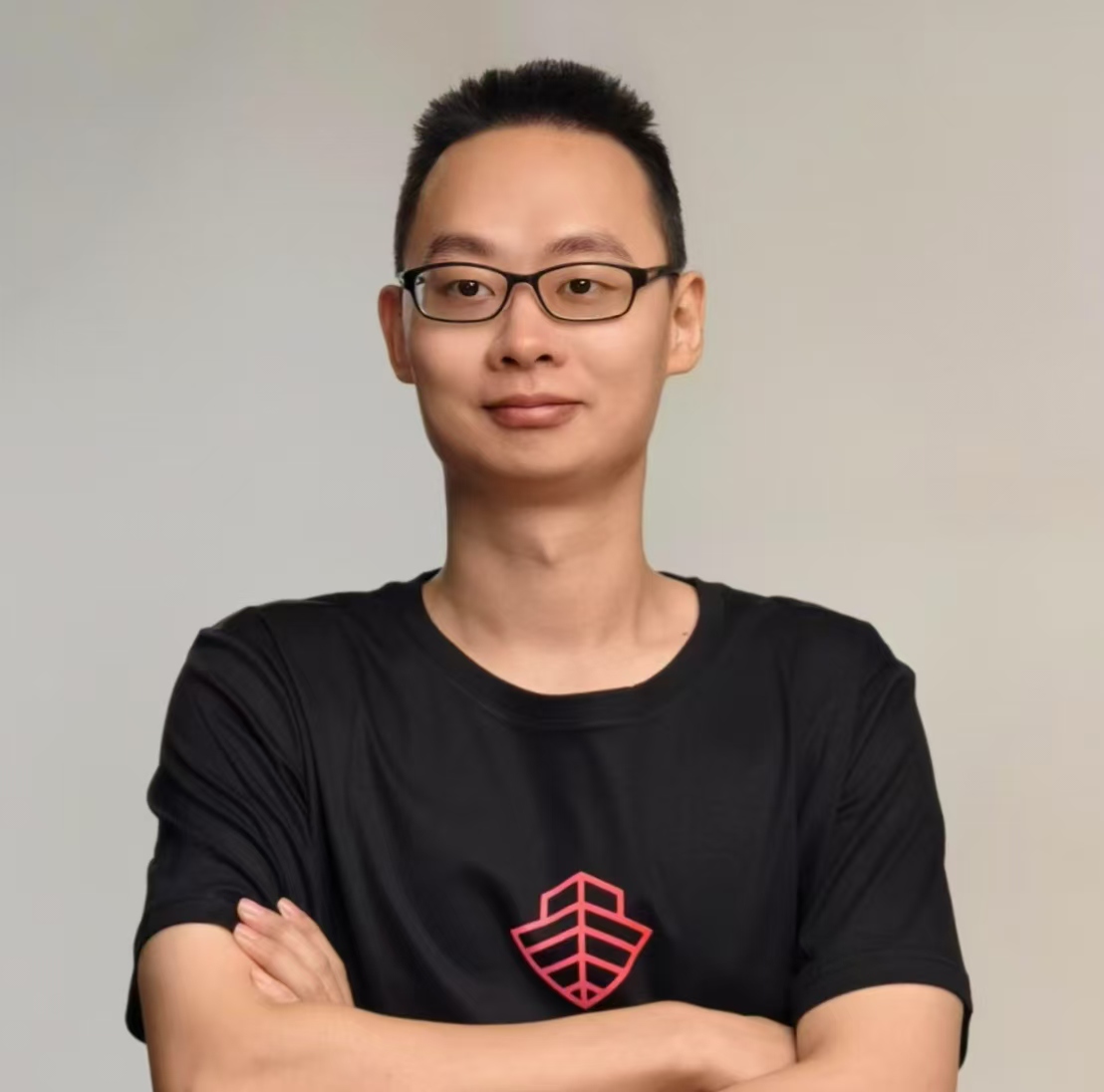}}]{Hang Xu} is currently a senior CV researcher with Yinwang Intelligent Technology Co., Ltd. He received his BSc from Fudan University and his Ph.D. from the University of Hong Kong. His research interests include multimodal large language models, autonomous driving, object detection, and AutoML. He has published over 100 papers at top AI conferences such as NeurIPS, CVPR, ICCV, AAAI.
\end{IEEEbiography}

\begin{IEEEbiography}[{\includegraphics[width=1in,height=1.25in,clip,keepaspectratio]{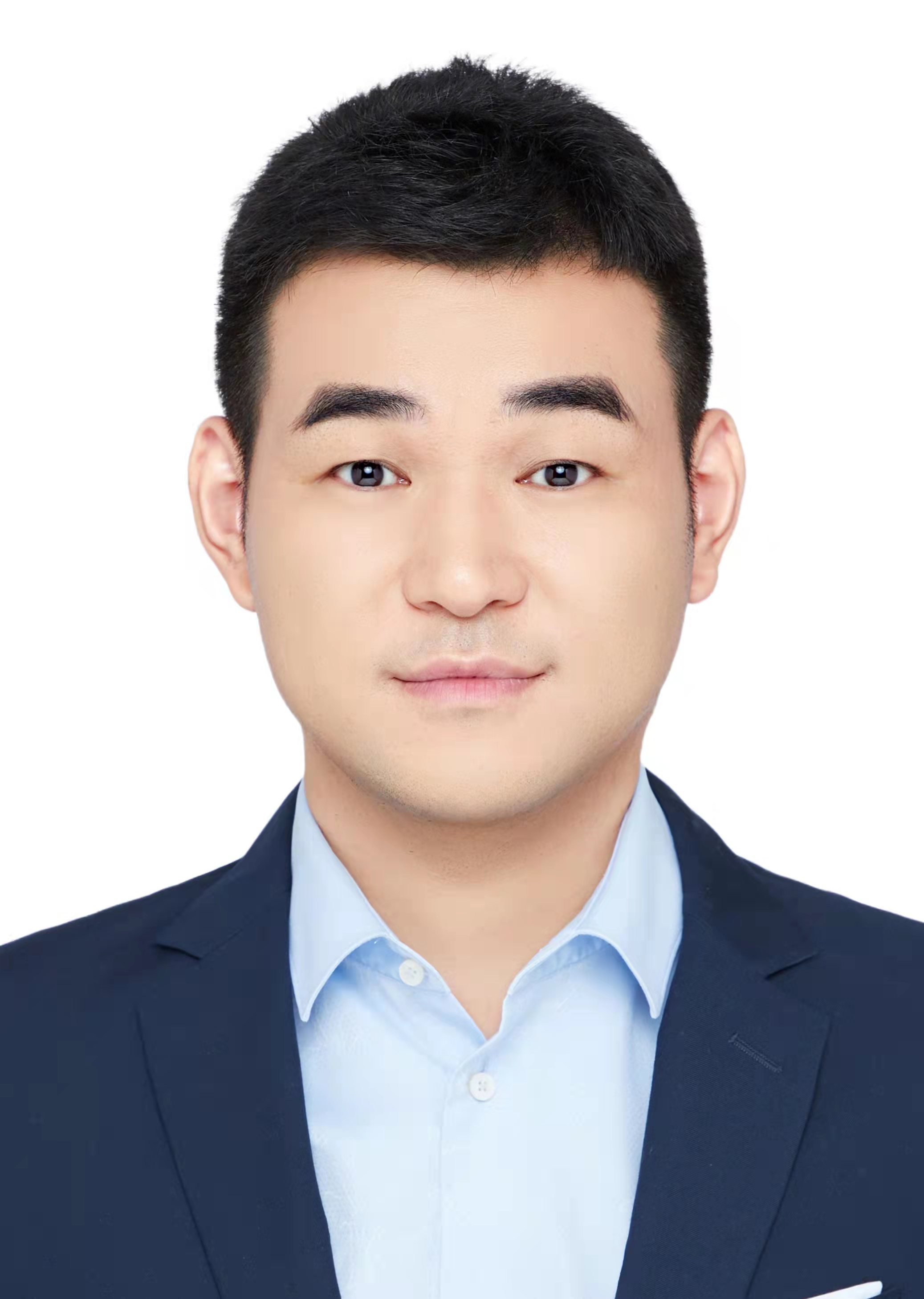}}]{Bokui Chen} received the Ph.D. degree from University of Science and Technology of China in 2013. He is currently an Assistant Professor at Tsinghua Shenzhen International Graduate School, Tsinghua University, China. His research interests include intelligent transportation systems and artificial intelligence.
\end{IEEEbiography}


\begin{IEEEbiography}[{\includegraphics[width=1in,height=1.25in,clip,keepaspectratio]{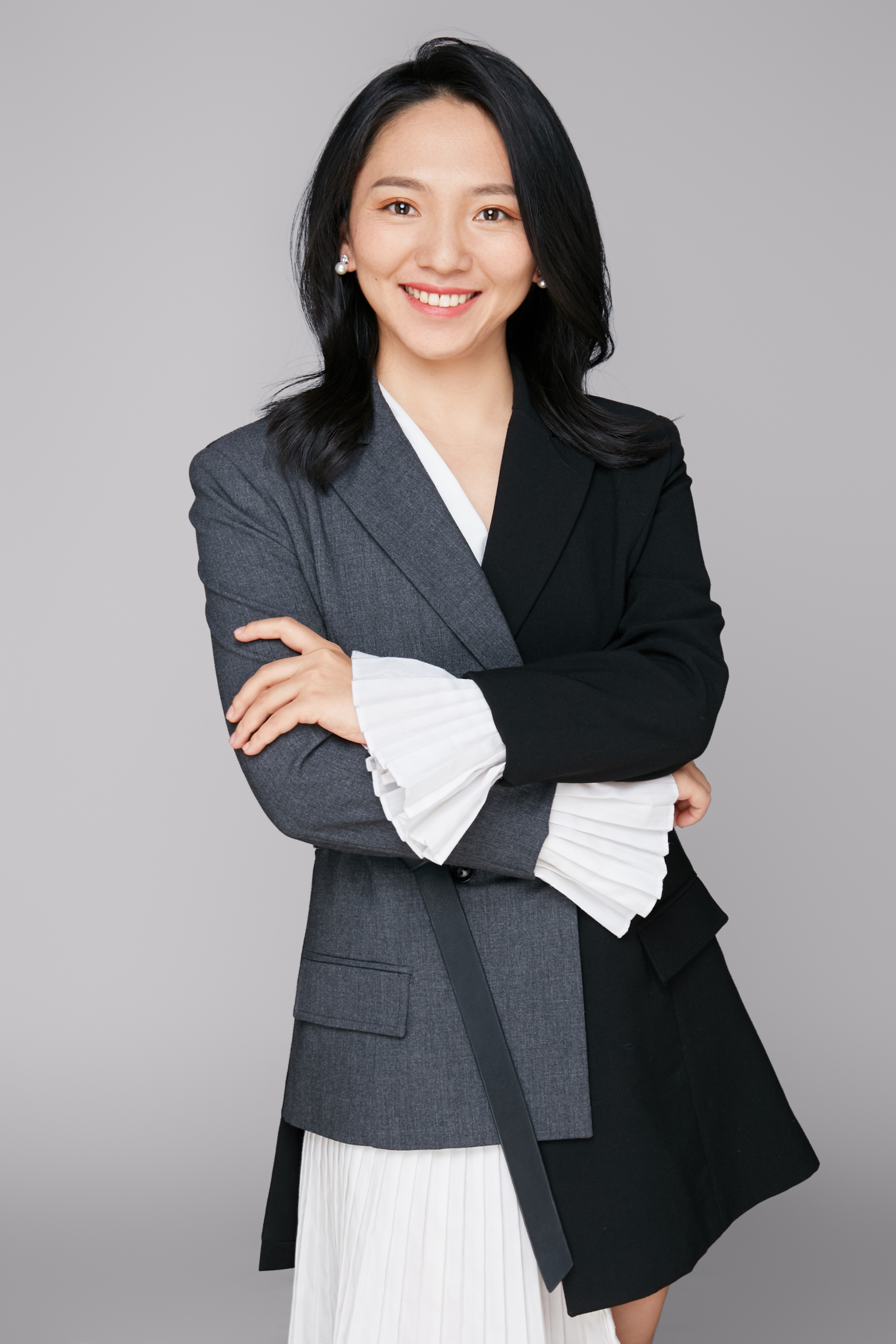}}]{Xiaodan Liang} 
received the Ph.D. degree from
Sun Yat-sen University, Guangzhou, China, in 2016,
advised by Liang Lin.
She was a Post-Doctoral Researcher with the
Machine Learning Department, Carnegie Mellon
University, Pittsburgh, PA, USA, from 2016 to 2018,
working with Prof. Eric Xing. She is currently
a Professor with Sun Yat-sen University. She has published several cutting-edge projects
on human-related analysis, including human parsing, pedestrian detection and instance segmentation,
2D/3D human pose estimation, and activity recognition.
\end{IEEEbiography}

\vfill

\end{document}